\documentclass{article}
\usepackage{ijcai21}

\usepackage{times}
\usepackage{soul}
\usepackage{url}
\usepackage[hidelinks]{hyperref}
\usepackage[utf8]{inputenc}
\usepackage[small]{caption}
\usepackage{graphicx}
\usepackage{amsmath}
\usepackage{amsthm}
\usepackage{booktabs}
\usepackage{fancyhdr}
\usepackage{amsmath}
\usepackage{booktabs}
\usepackage{threeparttable}
\usepackage{multirow}
\usepackage{arydshln}
\usepackage{subcaption}
\usepackage{array}
\usepackage{setspace}
\usepackage{algpseudocode}
\usepackage{bm}
\usepackage[inline]{enumitem}
\usepackage{amssymb}
\usepackage{soul}
\usepackage{tikz}
\usetikzlibrary{calc}
\usepackage{xparse}

\definecolor{orange2}{rgb}{0.77254902, 0.352941176, 0.066666667}
\definecolor{yellow2}{rgb}{0.749019608, 0.564705882, 0.0}
\definecolor{red}{rgb}{1.0,0.0,0.0}
\definecolor{orange}{rgb}{1.0,0.65,0.0}
\definecolor{green}{rgb}{0.0,0.75,0.0}
\definecolor{blue}{rgb}{0.0,0.0,1.0}
\definecolor{purple}{rgb}{0.5,0.0,0.5}

\newlength\LineWidth
\definecolor{HLcolor}{RGB}{124,18,18}
\sethlcolor{HLcolor}

\makeatletter

\newcommand{\defhighlighter}[3][]{%
  \tikzset{every highlighter/.style={draw=#2, fill opacity=#3, #1}}%
}

\defhighlighter[fill=olive!15]{HLcolor}{.25}

\newcommand{\highlight@DoHighlight}{
  \fill [outer sep = -15pt, inner sep = 0pt, every highlighter, this highlighter,draw=none]
        ($(begin highlight)+(0,8pt)$) rectangle ($(end highlight)+(0,-2pt)$) ;
  \draw[HLcolor,line width=\LineWidth]  ($(begin highlight)+(0,-2pt)$) -- ($(end highlight)+(0,-2pt)$) ;
  \draw[HLcolor,line width=\LineWidth]  ($(begin highlight)+(0,8pt)$) -- ($(end highlight)+(0,8pt)$) ;
}

\newcommand{\highlight@BeginHighlight}{
  \coordinate (begin highlight) at (0,0) ;
}

\newcommand{\highlight@EndHighlight}{
  \coordinate (end highlight) at (0,0) ;
}

\newdimen\highlight@previous
\newdimen\highlight@current

\DeclareRobustCommand*\highlight[1][]{%
  \tikzset{this highlighter/.style={#1}}%
  \SOUL@setup
  \def\SOUL@preamble{%
    \begin{tikzpicture}[overlay, remember picture]
      \highlight@BeginHighlight
      \draw[HLcolor,line width=\LineWidth]  ($(begin highlight)+(0,-2pt)+(0,-0.5\pgflinewidth)$) -- ($(begin highlight)+(0,8pt)+(0,0.5\pgflinewidth)$) ;
      \highlight@EndHighlight
    \end{tikzpicture}%
  }%
  \def\SOUL@postamble{%
    \begin{tikzpicture}[overlay, remember picture]
      \highlight@EndHighlight
      \highlight@DoHighlight
      \draw[HLcolor,line width=\LineWidth]  ($(end highlight)+(0,-2pt)+(0,-0.5\pgflinewidth)$) -- ($(end highlight)+(0,8pt)+(0,0.5\pgflinewidth)$) ;
    \end{tikzpicture}%
  }%
  \def\SOUL@everyhyphen{%
    \discretionary{%
      \SOUL@setkern\SOUL@hyphkern
      \SOUL@sethyphenchar
      \tikz[overlay, remember picture] \highlight@EndHighlight ;%
    }{%
    }{%
      \SOUL@setkern\SOUL@charkern
    }%
  }%
  \def\SOUL@everyexhyphen##1{%
    \SOUL@setkern\SOUL@hyphkern
    \hbox{##1}%
    \discretionary{%
      \tikz[overlay, remember picture] \highlight@EndHighlight ;%
    }{%
    }{%
      \SOUL@setkern\SOUL@charkern
    }%
  }%
  \def\SOUL@everysyllable{%
    \begin{tikzpicture}[overlay, remember picture]
      \path let \p0 = (begin highlight), \p1 = (0,0) in \pgfextra
        \global\highlight@previous=\y0
        \global\highlight@current =\y1
      \endpgfextra (0,0) ;
      \ifdim\highlight@current < \highlight@previous
        \highlight@DoHighlight
        \highlight@BeginHighlight
      \fi
    \end{tikzpicture}%
    \the\SOUL@syllable
    \tikz[overlay, remember picture] \highlight@EndHighlight ;%
  }%
  \SOUL@
}

\def\adl@drawiv#1#2#3{%
        \hskip.5\tabcolsep
        \xleaders#3{#2.5\@tempdimb #1{1}#2.5\@tempdimb}%
                #2\z@ plus1fil minus1fil\relax
        \hskip.5\tabcolsep}
\newcommand{\cdashlinelr}[1]{%
  \noalign{\vskip\aboverulesep
           \global\let\@dashdrawstore\adl@draw
           \global\let\adl@draw\adl@drawiv}
  \cdashline{#1}
  \noalign{\global\let\adl@draw\@dashdrawstore
           \vskip\belowrulesep}}

\makeatother

\DeclareDocumentCommand\MyDBox{O{HLcolor!15}O{HLcolor}m}{%
  \colorlet{HLcolor}{#1!75}
  \highlight[#1!75]{#3}%
}

\title{Interacting with Explanations through Critiquing}

\author{
    Diego Antognini$^1$\and Claudiu Musat$^2$\And Boi Faltings$^1$
\affiliations
$^1$École Polytechnique Fédérale de Lausanne, Switzerland\\
$^2$Swisscom, Switzerland
\emails
firstname.lastname@\{epfl.ch, swisscom.com\}
}

\begin{document}

\maketitle
\thispagestyle{fancy} 

\begin{abstract}

Using personalized explanations to support recommendations has been shown to increase trust and perceived quality. However, to actually obtain better recommendations, there needs to be a means~for users to modify the recommendation criteria by interacting with the explanation. We present~a novel explanation technique using aspect markers that learns to generate personalized explanations of recommendations from review texts, and we show that human users significantly prefer~these explanations over those produced by state-of-the-art techniques.

Our work's most important innovation is that it allows users to react to a recommendation by critiquing the textual explanation: removing (symmetrically adding) certain aspects they dislike or that are no longer relevant (symmetrically that are of interest). The system updates its user model~and~the resulting recommendations according to the critique. This is based on a novel unsupervised critiquing method for single- and multi-step critiquing with textual explanations. Empirical results show that our system achieves good performance in adapting to the preferences expressed in multi-step critiquing and generates consistent explanations.
\end{abstract}

\section{Introduction}
\textbf{Explanations of recommendations are beneficial}. Modern recommender systems accurately capture users' preferences and achieve high performance. But, their performance comes at the cost of increased complexity, which makes them seem like black boxes to users. This may result in distrust or rejection of the recommendations~\cite{tintarev2015explaining}.

There is thus value in providing \textit{textual explanations} of the recommendations, especially on e-commerce websites, because such explanations enable users to understand why~a particular item has been suggested and hence to make better decisions \cite{kunkel2018trust}. Furthermore, explanations increase overall system transparency \cite{tintarev2015explaining} and trustworthiness \cite{zhang2018exploring}. 

However, not all explanations are equivalent. \cite{kunkel2019let} showed that highly personalized justifications using \textit{natural language} lead to substantial increases in perceived recommendation quality and trustworthiness compared to simpler explanations, such as aspect, template, or similarity.

\begin{figure}[t]
\centering
\includegraphics[width=1.0\linewidth]{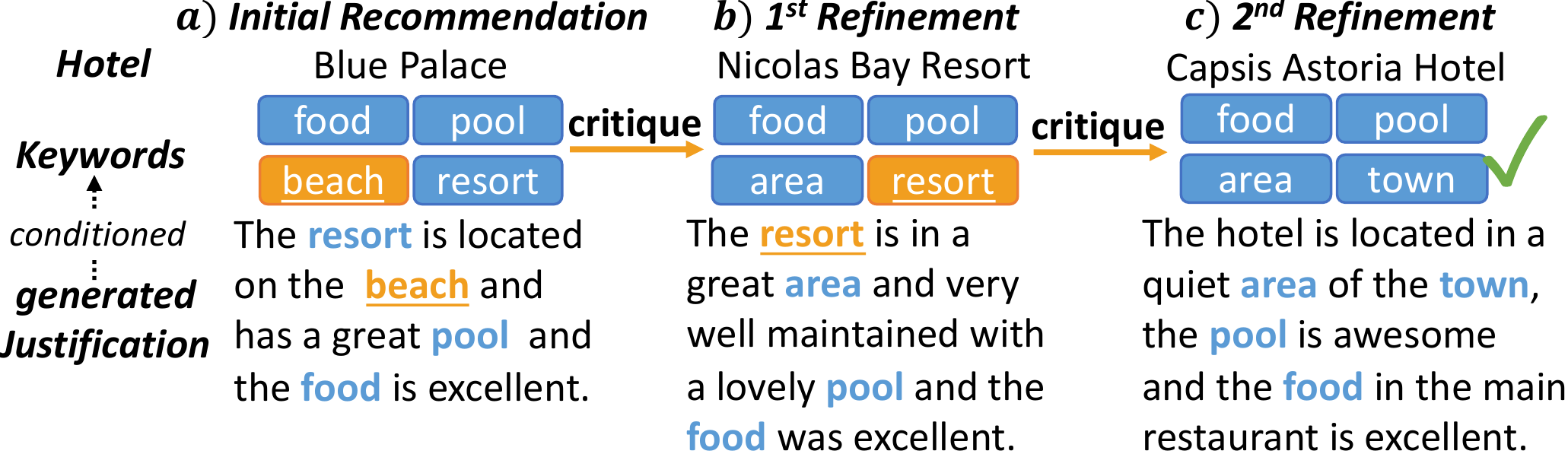}
\caption{\label{example_critiquing}A flow of conversational critiquing over two time steps.~a) The system proposes to the user a recommendation with a keyphrase explanation and a justification. The user can interact with the explanation and critique phrases. b) A new recommendation is produced from the user's profile and the critique. c) This process repeats until the user accepts the recommendation and ceases to provide critiques.}
\end{figure}

A second, and more important, benefit of explanations is that they provide a basis for feedback: if a~user~is unsatisfied~with~a recommendation, understanding what generated it allows them to \textit{critique} it (Fig.~\ref{example_critiquing}). Critiquing -- a~conversational method of incorporating user preference feedback regarding item attributes into the recommended list of items -- has several advantages. First, it allows the system to correct and improve an incomplete or inaccurate~model of the user's preferences, which improves the user's decision accuracy \cite{chen2012critiquing}. Compared to preference elicitation, critiquing is more flexible: users can express preferences in any order and on any criteria~\cite{reilly2005explaining}.

\textbf{Useful explanations are hard to generate}. Prior research has employed users' reviews to capture their preferences and writing styles (e.g., \cite{dong2017learning}). From past reviews, they generate \textit{synthetic} ones that serve as personalized \textit{explanations} of ratings given by users. However, many reviews are noisy, because they partly describe experiences or endorsements. It is thus nontrivial to identify meaningful justifications inside reviews. \cite{ni-etal-2019-justifying} proposed a pipeline for identifying justifications from reviews and asked humans to annotate them. \cite{chen2019co,chen2020towards} set the justification as the first sentence. However, these notions of justification were ambiguous,~and they assumed that a review contains only one justification.

\begin{figure}[t]
\centering
\includegraphics[width=0.811\linewidth]{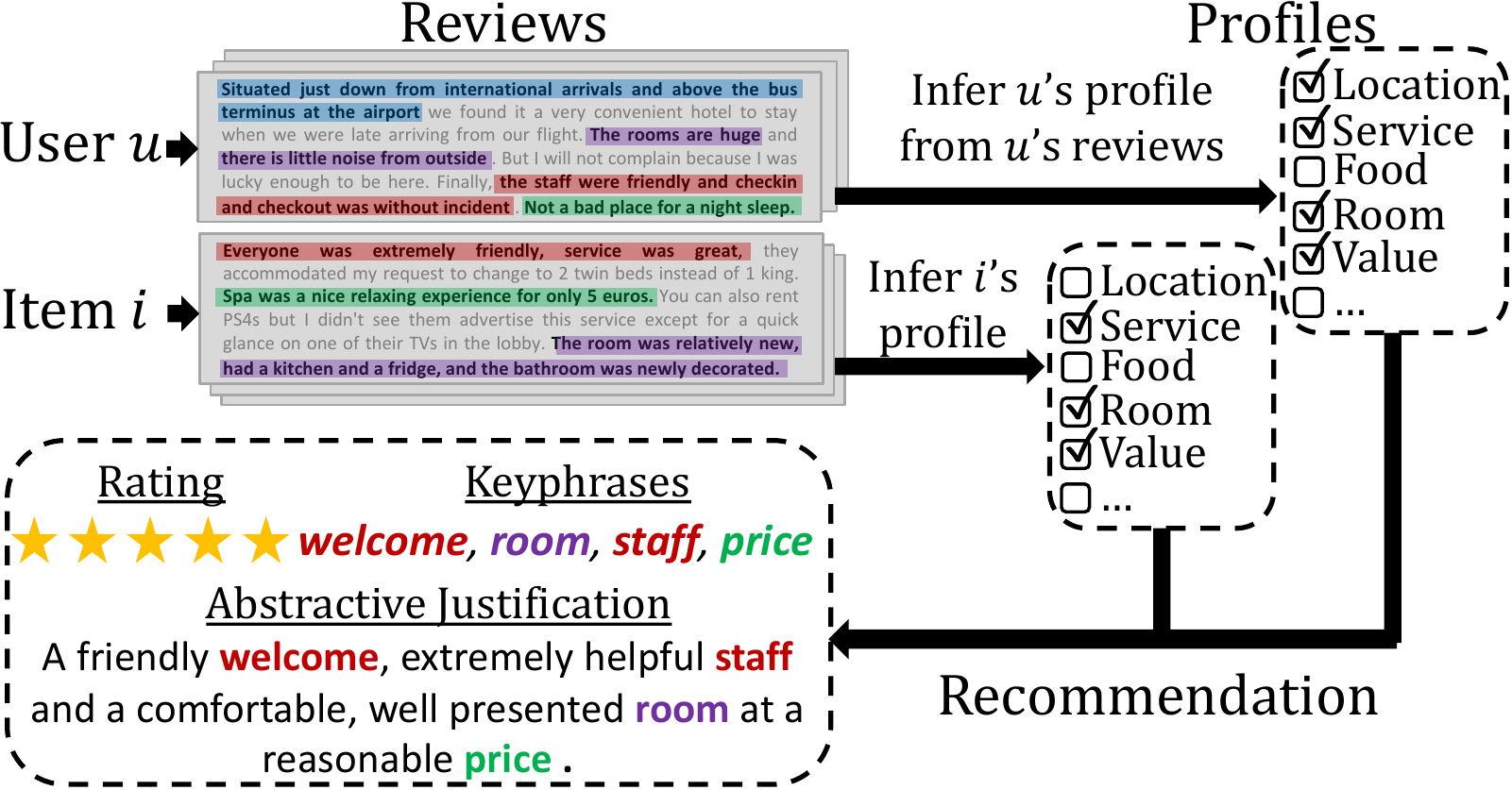}
\caption{\label{figure_pipeline}For reviews written by a user $u$ and a set of~reviews about an item $i$, we extract the justifications for each~aspect rating and implicitly build an interest profile. T-RECS outputs a personalized recommendation with~two~explanations: the keyphrases reflecting the overlap between the two profiles, and a synthetic justification conditioned on the~latter.}
\end{figure}

Recently, \cite{antognini2019multi} solved these shortcomings by introducing a justification extraction system with no prior limits imposed on their number or structure. This is important because a user typically justifies his overall rating with multiple explanations: one for each aspect the user cares about \cite{musat2015personalizing}. 
The authors showed that there is a connection between faceted ratings and snippets within the reviews: for each subrating, there exists at least one text fragment that alone suffices to make the prediction. They employed a sophisticated attention mechanism to favor long, meaningful word sequences; we call these \textbf{\textit{markers}}. Building upon their study, we show that these \textit{markers} serve to create better user and item profiles and can inform better user-item pair justifications. Fig.~\ref{figure_pipeline} illustrates the pipeline.

\textbf{From explanations to critiquing.} To reflect the overlap between the profiles of a user and an item, one can produce~a set of keyphrases and then a synthetic justification. The user can correct his profile, captured by the system, by \textit{critiquing} certain aspects he does not like or that are missing or not relevant anymore and obtain a new justification (Fig.~\ref{example_critiquing}). \cite{wu2019deep} introduced a keyphrase-based critiquing method in which attributes are mined~from reviews, and users interact with them. However, their models need~an~extra autoencoder to project the critique back into the latent space, and it is unclear how the models behave in multi-step~critiquing.

We overcome these drawbacks by casting the critiquing as an unsupervised attribute transfer task: altering a keyphrase explanation of a user-item pair representation to the critique. To this end, we entangle the user-item pair with the explanation in the same latent space. At inference, the keyphrase classifier modulates the latent representation until the classifier identifies it as the critique vector.

\textbf{In this work,} we address the problem recommendation with fine-grained explanations. We first demonstrate how~to extract multiple relevant and personalized justifications from the user's reviews to build a profile that reflects his preferences and writing style (Fig.~\ref{figure_pipeline}). Second, we propose T-RECS, a recommender with explanations. T-RECS explains a rating by first inferring a set of keyphrases describing the intersection between the profiles of a user and an item. Conditioned on the keyphrases, the model generates a synthetic personalized justification. We then leverage these explanations in an unsupervised critiquing method for single- and multi-step critiquing. We evaluate our model using two real-world recommendation datasets. T-RECS outperforms strong baselines in explanation generation, effectively re-ranks recommended items in single-step critiquing. Finally, T-RECS also better models the user's preferences in multi-step critiquing while generating consistent textual justifications.

\section{Related Work}
\textbf{Textual Explainable Recommendation}.
Researchers have investigated many approaches to generating textual explanations of recommended items for users. \cite{mcauley2013hidden} proposed a topic model to discover latent factors from reviews~and explain recommended items. \cite{zhang2014explicit} improved the understandability of topic words and aspects by filling template sentences. 

Another line of research has generated synthetic reviews as explanations. Prior studies have employed users' reviews and tips to capture their preferences and writing styles. \cite{catherine2017transnets} predicted and explained ratings by encoding the user's review and identifying similar~reviews. \cite{chen2019co} extended the previous work~to~generate short synthetic reviews. \cite{dual2020} optimized both tasks in dual forms. \cite{dong2017learning} proposed an attribute-to-sequence model to learn how to generate reviews given categorical attributes. \cite{ni2018personalized} improved review generation by leveraging aspect information using a seq-to-seq model with attention. Instead~of reviews, others have generated tips \cite{li2017neural,li2019persona}.
However, the tips are scarce and uninformative \cite{chen2019co}; many reviews are noisy because they describe partially general experiences or endorsements \cite{ni-etal-2019-justifying}. 

\cite{ni-etal-2019-justifying} built a seq-to-seq model conditioned on the aspects to generate relevant explanations for an existing recommender system;~the~fine-grained aspects are provided by the user in the inference. They identified justifications from reviews~by~segmenting them into elementary discourse units (EDU) \cite{mann1988rhetorical} and asking annotators to label them as ``good'' or ``bad'' justifications. \cite{chen2019co} set the justification as the first sentence. All assumed that a review contains only one justification. Whereas their notions of justification were ambiguous, we extract multiple justifications from reviews using \textit{markers} that justify subratings. Unlike their models, ours predicts keyphrases on which the justifications are conditioned and integrates critiquing.

\textbf{Critiquing}. Refining recommended items allows users~to interact with the system until they are satisfied. Some methods are example critiquing \cite{williams1982rabbit}, in which users critique a set of items; unit critiquing~\cite{unitcritiquing}, in which users critique an item's attribute and request another one instead; and compound critiquing \cite{reilly2005explaining} for more~aspects. The major drawback of these approaches is the assumption of a fixed set of known attributes.

\cite{wu2019deep} circumvented this limitation by extending the neural collaborative filtering model \cite{he2017neural}. First, the model explains a recommendation by predicting a set of keywords (mined from users' reviews). In \cite{chen2020towards}, based on \cite{chen2019co}, the model samples only one keyword via the Gumbel-Softmax function. Our work applies a deterministic strategy similar to \cite{wu2019deep}.

 Second, \cite{wu2019deep} project the critiqued keyphrase explanations back into the latent space, via an autoencoder that perturbs the training, from which the rating and the explanation are predicted. In this manner, the user's critique~modulates his latent representation. The model of \cite{chen2020towards} is trained in a two-stage manner: one to perform recommendation and predict one keyword and another to learn critiquing from online feedback, which requires additional data. By contrast, our model is simpler and learns critiquing in an~unsupervised fashion: it iteratively edits the latent representation until the new explanation matches the critique. 
 Finally, \cite{sannerwww20}~examined various linear aggregation methods on latent representations for multi-step critiquing. In comparison, our gradient-based critiquing iteratively updates the latent representation for each~critique.

\section{Extracting Justifications from Reviews}
\label{sec_extr_just}In this section, we introduce the pipeline for extracting high-quality and personalized justifications from users' reviews. We claim that a user justifies his overall experience with multiple explanations: one for each aspect he cares about. Indeed, it has been shown that users write opinions about the topics they care about \cite{zhang2014explicit}. Thus, the pipeline must satisfy two requirements:
\begin{enumerate*}
    \item extract text snippets that reflect a rating or subrating, and
    \item be data~driven and scalable to mine massive review corpora and to construct a large personalized recommendation justification dataset.
\end{enumerate*}
\begin{figure}[t]
\centering
\includegraphics[width=.772\linewidth]{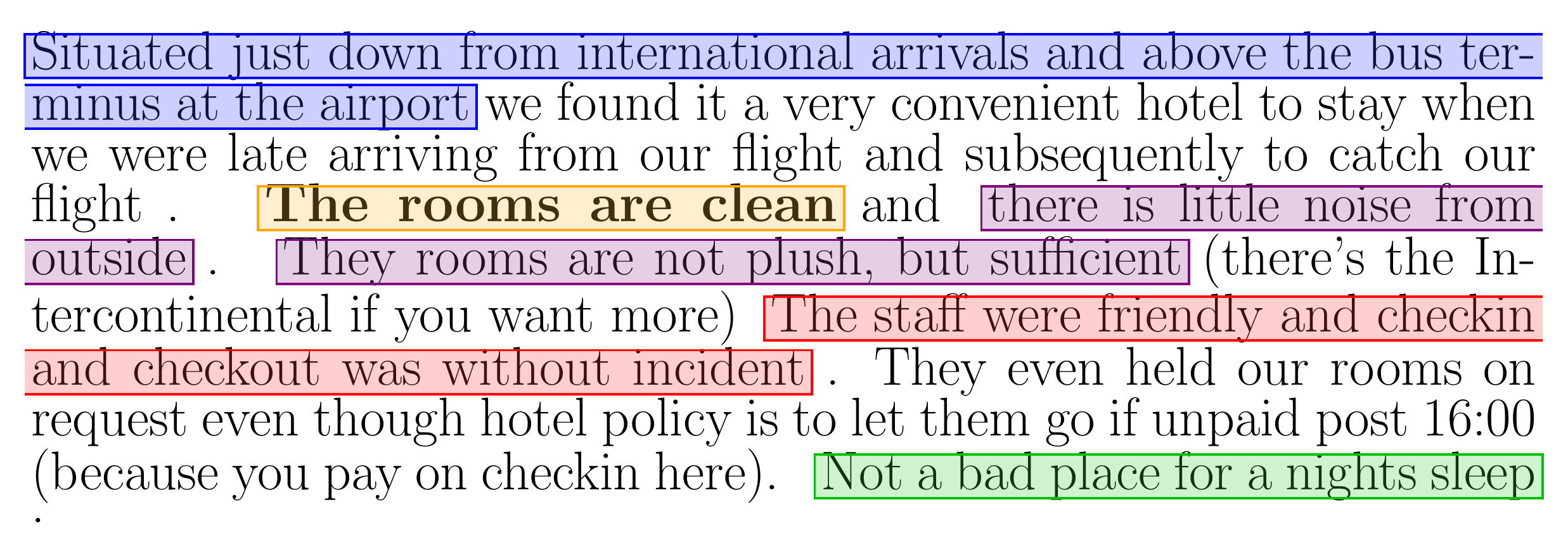}
\caption{\label{examples_masks_no_edus}Extracted justifications from a hotel review. The inferred \textit{markers} depict the excerpts that explain the ratings of the aspects: \MyDBox[red]{\strut Service}, \MyDBox[orange]{\strut Cleanliness}, \MyDBox[green]{\strut Value}, \MyDBox[purple]{\strut Room}, and \MyDBox[blue]{\strut Location}. We denote~in \textbf{bold} the EDU-based justification from the model of~\protect\cite{ni-etal-2019-justifying}.}
\end{figure}

\cite{antognini2019multi} proposed the multi-target masker (MTM) to find text fragments that explain faceted ratings in an unsupervised manner. MTM fulfills the two requirements. For each word, the model computes a distribution over the aspect set, which corresponds to the aspect ratings (e.g., service, location) and ``not aspect.'' In parallel, the model minimizes the number of selected words and discourages aspect transition between consecutive words. These two constraints guide the model to produce long, meaningful sequences of words called \textbf{\textit{markers}}. The model updates its parameters by using the inferred \textit{markers} to predict the aspect sentiments jointly and improves the quality of the \textit{markers} until convergence. 

Given a review, MTM extracts the \textit{markers} of each aspect. A sample is shown in Fig.~\ref{examples_masks_no_edus}. Similarly to \cite{ni-etal-2019-justifying}, we filter out \textit{markers}~that are unlikely to be suitable justifications: including third-person pronouns or being too short.~We use the constituency parse tree to ensure that \textit{markers} are verb phrases. The detailed processing is available in Appendix~\ref{app_process_filtering}.

\section{T-RECS: A Multi-Task Transformer with Explanations and Critiquing}
\label{sec_model}
\begin{figure*}[t]
\centering
   \includegraphics[width=0.83\linewidth]{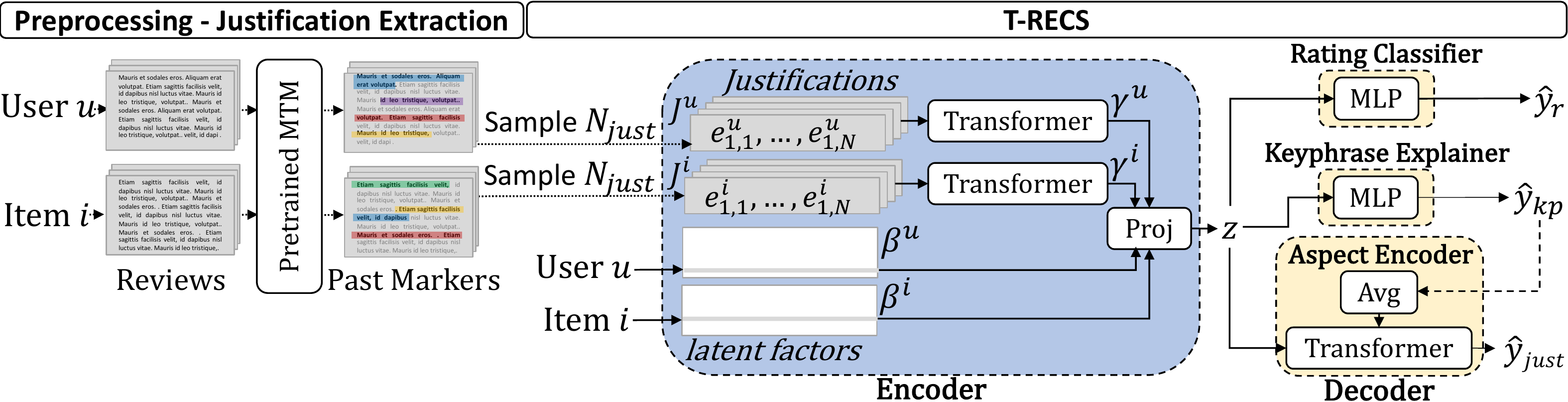}
   \caption{\label{general_architecture}(Left) Preprocessing for the users and the items.  
   For each user $u$ and item $i$, we first extract \textit{markers} from their past reviews (highlighted~in color), using the pretrained multi-target masker (see Section \ref{sec_extr_just}), that become their respective~justifications. Then, we sample $N_{just}$ of them and build the justification references $J^u$ and~$J^i$, respectively.
   (Right) T-RECS architecture. Given a user $u$ and an item $i$ with their justification references $J^u$, $J^i$ and latent factors $\beta^u$, $\beta^i$, T-RECS produces a joint embedding $\bm{z}$ from which it predicts a rating $\bm{\hat{y}}_r$, a keyphrase explanation $\bm{\hat{y}}_{kp}$, and a natural language justification $\bm{\hat{y}}_{just}$ conditioned on $\bm{\hat{y}}_{kp}$.}
\end{figure*}

Fig.~\ref{general_architecture} depicts the pipeline and our proposed T-RECS model. Let $U$ and $I$ be the user and item sets. For each user $u \in U$ (respectively an item $i \in I$), we extract \textit{markers} from the user's reviews on the training set, randomly select $N_{just}$, and build a justification~reference~$J^u$ (symmetrically~$J^i$). 

Given a user $u$, an item $i$, and their justification history $J^u$ and $J^i$, our goal is to~predict
\begin{enumerate*}
	\item a~rating~$\bm{y}_r$,
	\item a keyphrase explanation $\bm{y}_{kp}$ describing the relationship between $u$ and $i$, and 
	\item a natural language justification $\bm{y}_{just}=\{w_1, ..., w_N\}$
\end{enumerate*}, where $N$ is the length of the justification. $\bm{y}_{just}$ explains the rating $\bm{y}_r$ conditioned~on~$\bm{y}_{kp}$.

\subsection{Model Overview}
For each user and item, we extract \textit{markers} from their past reviews (in the train set) and build their justification history $J^u$ and $J^i$, respectively (see Section \ref{sec_extr_just}). T-RECS~is divided into four submodels: an \textbf{Encoder} $E$, which produces the latent representation $\bm{z}$ from the historical justifications and latent factors~of~the~user~$u$ and the item $i$; a \textbf{Rating Classifier} $C^r$, which classifies the rating $\bm{\hat{y}_r}$; a \textbf{Keyphrase Explainer} $C^{kp}$, which predicts~the keyphrase explanation $\bm{\hat{y}}_{kp}$ of the latent representation~$\bm{z}$; and a \textbf{Decoder}~$D$, which decodes the justification $\bm{\hat{y}}_{just}$ from~$\bm{z}$ conditioned on $\bm{\hat{y}}_{kp}$, encoded via the \textbf{Aspect Encoder} $A$. T-RECS involves four~functions: $\bm{z}=E(u,i)$;$\bm{\hat{y}}_r=C^r(\bm{z})$;$\bm{\hat{y}}_{kp} = C^{kp}(\bm{z})$;\ $\bm{\hat{y}}_{just} = D(\bm{z}, A(\bm{\hat{y}}_{kp})).$

The above formulation contains two types of personalized explanations: a list of keyphrases $\bm{\hat{y}}_{kp}$ that reflects the different aspects of item $i$ that the user $u$ cares about (i.e., the overlap between their profiles) and a natural language explanation~$\bm{\hat{y}}_{just}$ that justifies the rating, conditioned on $\bm{\hat{y}}_{kp}$. The set of keyphrases is mined from the reviews and reflects the different~aspects deemed important by the users. The keyphrases enable an interaction mechanism: users can express agreement or disagreement with respect to one or multiple aspects and hence critique the recommendation.

\subsubsection{Entangling user-item.} A key objective of T-RECS is to build a powerful latent representation. It accurately captures user and item profiles with their writing styles and entangles the rating, keyphrases, and~a natural language justification. Inspired by the superiority of the Transformer for text generation~tasks~\cite{radford2019language}, we propose a Transformer-based encoder that learns latent personalized features from users' and items' justifications. We first pass each justification $J^u_j$ (respectively $J^i_j$) through the Transformer to compute the intermediate representations $\bm{h}^u_j$ (respectively $\bm{h}^i_j$). We apply a sigmoid function on the representations and average them to get $\bm{\gamma}^u$ and $\bm{\gamma}^i$:\begin{equation*}
    \bm{\gamma}^u = \frac{1}{|J^u|}\sum\nolimits_{j \in J^u} \sigma(\bm{h}^u_j) \quad \bm{\gamma}^i = \frac{1}{|J^i|}\sum\nolimits_{j \in J^i} \sigma(\bm{h}^i_j).
\end{equation*}In parallel, the encoder maps the user $u$ (item $i$) to the latent factors $\bm{\beta}^u$ ($\bm{\beta}^i$) via an embedding layer. We compute~the~latent representation $\bm{z}$ by concatenating the latent personalized features and factors and applying a linear projection:
~$\bm{z}=$ $E(u,i)= W[\bm{\gamma}^u \mathbin\Vert \bm{\gamma}^i \mathbin\Vert \bm{\beta}^u \mathbin\Vert \bm{\beta}^i] + \bm{b},$
where~$\mathbin\Vert$~is the concatenation operator, and $W$ and $\bm{b}$ the projection parameters.

\subsubsection{Rating Classifier \& Keyphrase Explainer.} Our framework classifies the interaction between the user~$u$ and item $i$ as positive or negative. Moreover, we predict the keyphrases that describe the overlap of their profiles.~Both models are a two-layer feedforward neural network with LeakyRelu activation function. Their respective losses~are:\begin{equation*}
\begin{split}
    \mathcal{L}_{r}(C^r(\bm{z}),\bm{y}_r) &= (\bm{\hat{y}}_r - \bm{y}_r)^2\\
    \mathcal{L}_{kp}(C^{kp}(\bm{z}),\bm{y}_{kp}) &= -\sum\nolimits_{k=1}^{|K|} y_{kp}^k \log \hat{y}_{kp}^k
\end{split}
\end{equation*}where $\mathcal{L}_{r}$ is the mean square error, $\mathcal{L}_{kp}$ the binary cross-entropy, and $K$ the whole set of keyphrases.
\subsubsection{Justification Generation.} The last component consists of generating the justification. Inspired by ``plan-and-write''~\cite{yao2019plan}, we advance the personalization of the justification by incorporating the keyphrases $\bm{\hat{y}}_{kp}$. In other words, T-RECS generates a natural language justification conditioned on the
\begin{enumerate*}
    \item user,
    \item item, and
    \item aspects of the item that the user would consider important.
\end{enumerate*}
We encode these via the Aspect Encoder $A$ that takes the average of their word embeddings from the embedding layer in the Transformer. The aspect embedding is denoted by $\bm{a}_{kp}$ and added to the latent representation: $\bm{\tilde{z}} = \bm{z} + \bm{a}_{kp}$. Based on $\bm{\tilde{z}}$, the Transformer decoding block computes the output probability $\hat{y}^{t,w}_{just}$ for the word $w$ at time-step $t$. We train using teacher-forcing and cross-entropy with label smoothing:\begin{equation*}
 \mathcal{L}_{just}(D(\bm{z}, \bm{a}_{kp}), \bm{y}_{just}) = - \sum\nolimits_{t=1}^{|\bm{y}_{just}|} CE(y^{t,w}_{just}, \hat{y}^{t,w}_{just})
\end{equation*}We train T-RECS end-to-end and minimize jointly~the~loss $\bm{\mathcal{L}} = \lambda_{r} \mathcal{L}_{r} + \lambda_{kp} \mathcal{L}_{kp} + \lambda_{just} \mathcal{L}_{just}$,~where $\lambda_{r}$, $\lambda_{kp}$, and $\lambda_{just}$ control the impact of each loss. All objectives share the latent representation~$\bm{z}$ and are thus mutually regularized by the function $E(u,i)$ to limit overfitting by any~objective.
\subsection{Unsupervised Critiquing}
\label{sec_critiquing}
The purpose of critiquing is to refine the recommendation based on the user's interaction with the explanation, the keyphrases $\bm{\hat{y}}_{kp}$, represented with a binary vector. The user critiques either a keyphrase $k$ by setting $\hat{y}_{kp}^k=0$ (i.e., disagreement) or symmetrically adding a new one (i.e.,$\hat{y}_{kp}^k=1$). We denote the critiqued keyphrase explanation as~$\bm{\tilde{y}}^*_{kp}$.\begin{figure}[t]
\centering
   \includegraphics[width=0.853\linewidth]{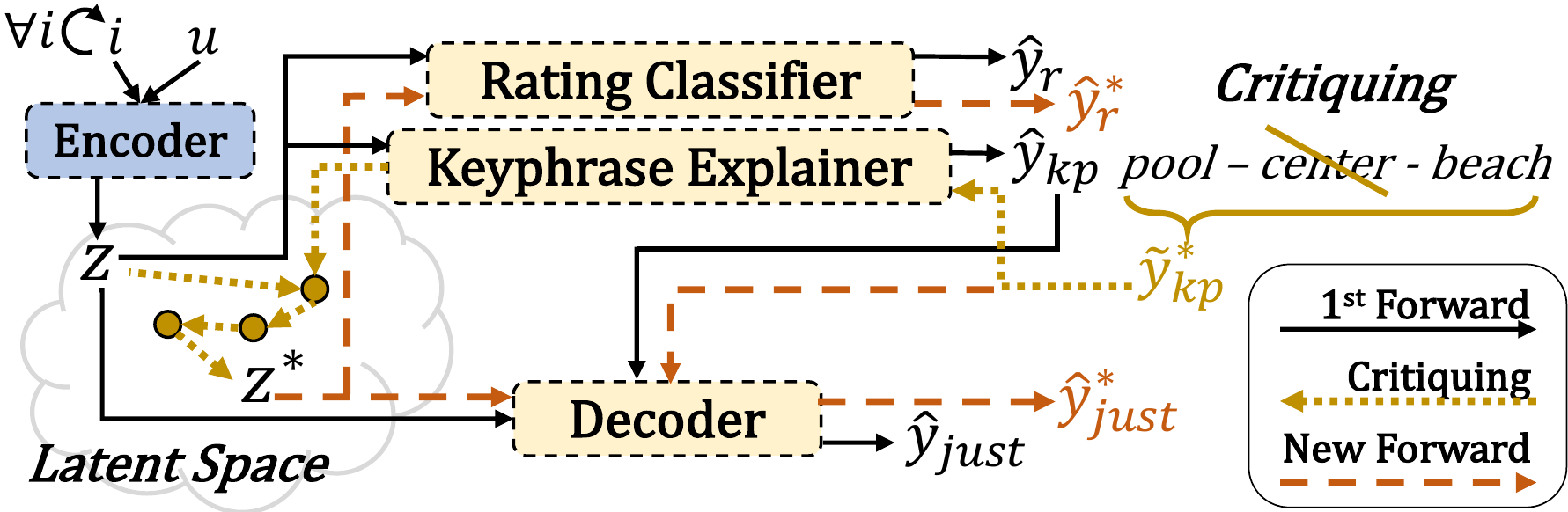}
   \caption{\label{general_critiquing}Workflow of considering to recommend items to~a user~$u$. We illustrate it for a given item $i$. \textbf{Black} denotes~the forward pass to infer the rating $\bm{\hat{y}}_r$ with the explanations $\bm{\hat{y}}_{kp}$ and $\bm{\hat{y}}_{just}$.~\textcolor{yellow2}{\textbf{Yellow}} indicates the critiquing: the user critiques the binary-vector keyphrase explanation $\bm{\hat{y}}_{kp}$ (e.g., \textit{center}) to $\bm{\tilde{y}^*}_{kp}$, which modulates the latent space into $\bm{z}^*$ for each item. \textcolor{orange2}{\textbf{Orange}} shows the new forward pass for the subsequent recommendation $\bm{\hat{y}}^*_r$ and explanations $\bm{\hat{y}}^*_{kp}$, $\bm{\hat{y}}^*_{just}$.}
\end{figure} The overall critiquing process is depicted in Fig.~\ref{general_critiquing}. Inspired by the recent success in editing the latent space on the unsupervised text attribute transfer task~\cite{abs-1905-12926}, we employ the trained Keyphrase Explainer $C^{kp}$ and the critiqued explanation $\bm{\tilde{y}}^*_{kp}$ to provide the gradient from which we update the latent representation $\bm{z}$ (depicted in \textcolor{yellow2}{\textbf{yellow}}). More formally, given a latent representation $\bm{z}$ and a binary critique vector $\bm{\tilde{y}}^*_{kp}$, we want to find a new latent representation $\bm{z}^*$ that will produce a new keyphrase explanation close to the critique, such that $|C^{kp}(\bm{z}^*) - \bm{\tilde{y}}^*_{kp}| \le T$, where $T$ is a threshold. In order to achieve this goal, we iteratively compute the gradient \underline{with respect to $\bm{z}$ instead of the model parameters $C^{kp}_\theta$}. We then modify $\bm{z}$ in the direction of the gradient until we get a new latent representation $\bm{z}^*$ that $C^{kp}$ considers close enough to~$\bm{\tilde{y}}^*_{kp}$ (shown in \textcolor{orange2}{\textbf{orange}}). We emphasize that we use the gradient to modulate $\bm{z}$ rather than the parameters~$C^{kp}$. 

Let denote the gradient as $\bm{g}_t$ and a decay coefficient as $\zeta$. For each iteration~$t$ and $\bm{z}^*_0 = \bm{z}$, the modified latent representation $\bm{z}^*_t$ at the $t$\textsuperscript{th} iteration can be formulated as follows:\begin{equation*}
    \bm{g}_t = \nabla_{\bm{z}^*_t} \mathcal{L}_{kp}(C^{kp}(\bm{z}^*_t), \bm{\tilde{y}}^*_{kp}); \ \ \bm{z}^*_t = \bm{z}^*_{t-1} - \zeta^{t-1} \bm{g}_t/||\bm{g}_t||_2
\end{equation*}Because this optimization is nonconvex, there is no guarantee that the difference between the critique vector and the inferred explanation will differ by only $T$. In our experiments in Section~\ref{sec_rq4}, we found that a limit of $50$ iterations works well, and that the newly induced explanations remain consistent.

\section{Experiments}

\subsection{Experimental Settings}

\paragraph{Datasets.} We evaluate the quantitative performance of T-RECS using two real-world, publicly available datasets: BeerAdvocate~\cite{mcauley2013hidden} and HotelRec~\cite{antognini_hotel_rec}. They contain 1.5 and~50 million reviews from BeerAdvocate and TripAdvisor. In addition to the overall rating, users also provided five-star aspect ratings. 
We binarize the ratings with a threshold $t$: $t > 4$~for hotel reviews and $t > 3.5$ for beer reviews. We further filter out all users with fewer than 20 interactions and sort them chronologically. We keep the first~80\%~of interactions per user as the training data, leaving the remaining 20\% for validation and testing. We sample two justifications per review. 

We need to select keyphrases for explanations and critiquing. Hence, we follow the processing in~\cite{wu2019deep}~to~extract 200 keyphrases (distributed uniformly over the aspect categories) from the \textit{markers} on each dataset. 

\paragraph{Implementation Details.} To extract \textit{markers}, we trained MTM with the hyperparameters reported by the authors. 
We build the justification history $J^u$,$J^i$, with $N_{just}=32$. We set the embedding and attention dimension to 256 and to 1024 for the~feed-forward network. The encoder and decoder consist of two layers of Transformer with $4$~attention heads. We use a batch size of 128, dropout of 0.1, and Adam with learning rate $0.001$.
For critiquing, we choose a threshold and decay coefficient $T=0.015,\zeta=0.9$ and $T=0.01,\zeta=0.975$~for hotel and beer reviews. We tune all models on the dev set. For reproducibility purposes, we provide details~in~Appendix~\ref{app_reproducibility}.

\subsection{RQ 1: Are \textit{markers} appropriate justifications?}
\label{sec_rq1}
We derive baselines from \cite{ni-etal-2019-justifying}: we split a review into elementary discourse~units (EDUs) and apply their classifier to get justifications; it is trained on~a manually~annotated dataset and generalizes well to other domains.~We employ two variants: EDU One and EDU All. The latter~includes all justifications, whereas the former includes only one.

We perform a human evaluation using Amazon's Mechanical Turk (see Appendix~\ref{app_hes} for more details) to judge the quality of the justifications extracted from the Markers, EDU One, and EDU All on both datasets. We employ three~setups: an evaluator is presented with 1. the three types of justifications; 2. only those from Markers and EDU All; and~3. EDU One instead of EDU All. We sampled 300 reviews (100 per setup) with generated justifications presented in random order. The annotators judged the justifications by choosing the most convincing in the pairwise setups and otherwise using best-worst scaling. We report the win rates for the pairwise comparisons and a normalized score ranging from -1 to +1.

Table~\ref{he_which_justification} shows that justifications extracted from Markers are preferred, on both datasets, more than 80\% of the time. Moreover, when compared to EDU All and EDU One, Markers achieve a score of $0.74$, three times higher than EDU All. Therefore, justifications extracted from the Markers are significantly better than EDUs, and a single justification cannot explain a review. Fig.~\ref{examples_masks_no_edus} shows a sample for comparison.

\subsection{RQ 2: Does T-RECS generate high-quality, relevant, and personalized explanations?}
\begin{table}[t]
\small
    \centering
   \caption{\label{stats_datasets}Descriptive statistics of the datasets.}
\begin{threeparttable}
\begin{tabular}{@{}l@{\hspace{0.5mm}}c@{\hspace{1mm}}c@{\hspace{1mm}}c@{\hspace{1mm}}c@{\hspace{1mm}}c@{\hspace{1mm}}c@{\hspace{1mm}}c@{\hspace{1mm}}c@{}}
& & & & & & \multicolumn{3}{c}{Avg. \#KP per}\\
\cmidrule{7-9}
\textbf{Dataset} & \textbf{\#Users} & \textbf{\#Items} & \textbf{\#Inter.} & \textbf{Dens.} & \textbf{KP Cov.} & \textbf{Just.} & \textbf{Rev.} & \textbf{User}\\
\toprule
Hotel & 72,603 & 38\,896& 2.2M & 0.08\% & 97.66\% & 2.15& 3.79 & 115\\
Beer & 7,304 & 8,702 & 1.2M & 2.02\% & 96.87\% & 3.72 & 6.97 & 1,210\\
\end{tabular}
\end{threeparttable}
\end{table}
\begin{table}[t]
\small
    \centering
   \caption{\label{he_which_justification}Human evaluation of explanations in terms of the win rate and the \textbf{b}est-\textbf{w}orst scaling. A score significantly different than Markers (post hoc Tukey HSD test) is denoted by {*}{*} for $p < 0.001$.}
\begin{threeparttable}
\begin{tabular}{@{}l@{}c@{}c@{\hspace{2mm}}c@{\hspace{1mm}}c@{\hspace{3mm}}c@{}c@{\hspace{2mm}}c@{}}
& \multicolumn{3}{c}{\textit{Hotel}} & & \multicolumn{3}{c}{\textit{Beer}}\\
\cmidrule{2-4}\cmidrule{6-8}
\textbf{Winner} \textbf{Loser} & \multicolumn{3}{c}{\textbf{Win Rate}} & &  \multicolumn{3}{c}{\textbf{Win Rate}}\\
\cmidrule[0.08em]{1-8}
Markers EDU All & & 81\%{*}{*} & & & & 77\%{*}{*} & \\
Markers EDU One & & 93\%{*}{*} & & & & 90\%{*}{*} & \\
\cdashlinelr{1-8}
\textbf{Model} & \textbf{Score} & \textbf{\#B} & \textbf{\#W} & & \textbf{Score} & \textbf{\#B} & \textbf{\#W}\\
\cmidrule[0.08em]{1-8}
EDU One & -0.95{*}{*} & \ 1 & 96 & & -0.93{*}{*} & \ 2 & 95\\
EDU All & \ 0.21{*}{*} & 24 & \ 3 & & \ 0.20{*}{*} & 23 & \ 3\\
Markers & \textbf{0.74\ \ } & 75 & \ 1 & & \textbf{0.73}\ \  & 75 & \ 2\\
\end{tabular}
\end{threeparttable}
\end{table} 
\label{sec_rq2}
\paragraph{Natural Language Explanations.}
\label{sec_nle_justifications}We consider five baselines: 
ExpansionNet~\cite{ni2018personalized} is a seq-to-seq model with a user, item, aspect, and fusion attention mechanism that generates personalized reviews. DualPC~\cite{dual2020} and CAML~\cite{chen2019co} generate an explanation based on a rating and the user-item pair. Ref2Seq improves upon ExpansionNet by learning only from historical justifications of a user and an item. AP-Ref2Seq~\cite{ni-etal-2019-justifying} extends Ref2Seq with aspect planning~\cite{yao2019plan}, in~which aspects are given during the generation. All models use~beam search during testing and the same keyphrases~as~aspects.
We employ BLEU, ROUGE-L, BertScore~\cite{bert-score}, the perplexity for the fluency, and R\textsubscript{KW} for the explanation consistency as in \cite{chen2020towards}: the ratio of the target keyphrases present in the generated justifications.

The main results are presented in Table~\ref{just_auto_perfs} (more in Appendix~\ref{app_all_results}).~T-RECS achieves the highest scores on both datasets. 
We note that
\begin{enumerate*}
    \item seq-to-seq models better capture user and item information to produce more relevant justifications, and 
    \item using a keyphrase plan doubles the performance on average and improving explanation consistency.
\end{enumerate*}

We run a human evaluation, with the best models according to $R_{KW}$, using best-worst scaling on the dimensions:~overall, fluency, informativeness, and relevance.
We sample 300 explanations and showed them in random order. Table~\ref{just_human_perfs} shows that our explanations are largely preferred on all criteria. 

\paragraph{Keyphrase Explanations.}
\begin{table}[t]
\small
    \centering
\caption{\label{just_auto_perfs}Generated justifications on automatic evaluation.}
\begin{threeparttable}
\begin{tabular}{@{}cl@{}c@{\hspace*{2mm}}c@{\hspace*{1mm}}c@{\hspace*{2mm}}c@{\hspace*{4mm}}c@{}}
& \textbf{Model} &  \textbf{BLEU} & \textbf{R-L} & \textbf{BERT\textsubscript{Score}} & \textbf{PPL}$\downarrow$ & \textbf{R\textsubscript{KW}}\\
\toprule
\multirow{6}{*}{\rotatebox{90}{\textit{Hotel}}}
& ExpansionNet & 0.53 & 6.91 & 74.81 & 28.87 & 60.09\\
& DualPC & 1.53 & 16.73 & 86.76 & 28.99 & 13.12\\
& CAML & 1.13 & 16.67 & 87.77 & 29.10 & 9.38\\
& Ref2Seq & 1.77 & 16.45 & 86.74 & 29.07 & 13.19\\
& AP-Ref2Seq & 7.28 & 33.71 & 88.31 & 21.31 & 90.20\\
& T-RECS & \textbf{7.47} & \textbf{34.10} & \textbf{90.23} & \textbf{17.80} & \textbf{93.57}\\
\midrule
\multirow{6}{*}{\rotatebox{90}{\textit{Beer}}}
& ExpansionNet & 1.22 & 9.68 & 72.32 & 22.28 & 82.49\\
& DualPC & 2.08 & 14.68 & 85.49 &21.15 & 10.60\\
& CAML & 2.43 & 14.99 & 85.96 & 21.29& 10.18\\
& Ref2Seq & 3.51 & 15.96 & 85.27 & 22.34 & 12.10\\
& AP-Ref2Seq & 15.89 & 46.50 & 91.35 & 12.07 & 91.52\\
& T-RECS & \textbf{16.54} & \textbf{47.20} & \textbf{91.50} & \textbf{10.24} & \textbf{94.96}\\
\end{tabular}
\end{threeparttable}
\end{table}\begin{table}[t]
\small
    \centering
   \caption{\label{just_human_perfs}Human evaluation of justifications in terms of best-worst scaling for \textbf{O}verall, \textbf{F}luency, \textbf{I}nformativenss, and \textbf{R}elevance. Most scores are significantly different than T-RECS (post hoc Tukey HSD test) with $p < 0.002$. $\dagger$ denotes a nonsignificant score.}
\begin{threeparttable}
\begin{tabular}{@{}l@{\hspace*{1mm}}c@{\hspace*{1.5mm}}c@{\hspace*{1.5mm}}c@{\hspace*{1.5mm}}c@{}c@{\hspace*{3mm}}c@{\hspace*{1.5mm}}c@{\hspace*{1.5mm}}c@{\hspace*{1.5mm}}c@{}}
& \multicolumn{4}{c}{\textit{Hotel}} & & \multicolumn{4}{c}{\textit{Beer}}\\
\cmidrule{2-5}\cmidrule{7-10}
\textbf{Model} & \textbf{O} & \textbf{F} & \textbf{I} & \textbf{R} & & \textbf{O} & \textbf{F} & \textbf{I} & \textbf{R}\\
\toprule
ExpansionNet & -0.58 & -0.67 & -0.52 & -0.56 & & -0.03 & -0.31 & 0.10 & -0.01\\
Ref2Seq & -0.27 & -0.19 & -0.30 & -0.26 & & -0.69 & -0.34 & -0.71 & -0.69\\
AP-Ref2Seq & 0.30 & 0.32 & 0.29 & 0.29 & & 0.22 & 0.25$\dagger$ & 0.21$\dagger$ & 0.25\\
T-RECS & \textbf{0.55} & \textbf{0.54} & \textbf{0.53} & \textbf{0.53} & & \textbf{0.49} & \textbf{0.39} & \textbf{0.39} & \textbf{0.45}\\
\end{tabular}
\end{threeparttable}
\end{table}
\label{sec_keyphrases_explanations}
We compare T-RECS with the popularity baseline and the models proposed in~\cite{wu2019deep}, which are extended versions~of the NCF model~\cite{he2017neural}. E-NCF and CE-NCF augment the NCF method with an \underline{e}xplanation and a~\underline{c}ritiquing neural component. Also, the authors provide \underline{v}ariational variants: VNCF, E-VNCF, and CE-VNCF. Here, we omit NCF and VNCF because they are trained only to predict ratings. We report the following metrics: NDCG, MAP, Precision, and Recall at~10.\begin{table}[t]
\small
    \centering
\caption{\label{kw_perfs}Keyphrase explanation quality at $N=10$.}
\begin{threeparttable}
\begin{tabular}{@{}l@{}c@{\hspace{1mm}}c@{\hspace{1.5mm}}c@{\hspace{1.5mm}}c@{}c@{\hspace{2mm}}c@{\hspace{1mm}}c@{\hspace{1.5mm}}c@{\hspace{1.5mm}}c@{}}
& \multicolumn{4}{c}{\textit{Hotel}} & & \multicolumn{4}{c}{\textit{Beer}}\\
\cmidrule{2-5}\cmidrule{7-10}
\textbf{Model} & \textbf{NDCG} & \textbf{MAP} & \textbf{P} & \textbf{R} &  & \textbf{NDCG} & \textbf{MAP} & \textbf{P} & \textbf{R}\\
\toprule
Pop & 0.333 & 0.208 & 0.143 & 0.396 &  & 0.250 & 0.229 & 0.176 & 0.253\\
E-NCF & 0.341 & 0.215 & 0.137 & 0.380 &  & 0.249 & 0.220 & 0.179 & 0.262\\
CE-NCF & 0.229 & 0.143 & 0.092 & 0.255 &  & 0.192 & 0.172 & 0.136 & 0.197\\
E-VNCF & 0.344 & 0.216 & 0.139 & 0.386 &  & 0.236 & 0.210 & 0.170 & 0.248\\
CE-VNCF & 0.229 & 0.134 & 0.107 & 0.297 &  & 0.203 & 0.178 & 0.148 & 0.215\\
T-RECS & \textbf{0.376} & \textbf{0.236} & \textbf{0.158} & \textbf{0.436} &  & \textbf{0.316} & \textbf{0.280} & \textbf{0.228} & \textbf{0.332}\\
\end{tabular}
\end{threeparttable}
\end{table}

Table~\ref{kw_perfs} shows that T-RECS outperforms the CE-(V)NCF models by 60\%, Pop by 20\%, and E-(V)NCF models by 10\% to 30\% on all datasets. Pop performs better than~CE-(V)NCF, showing that many keywords are recurrent in reviews. Thus, predicting keyphrases from the user-item latent space is a natural~way to entangle them with (and enable~critiquing). 

\subsection{RQ 3: Can T-RECS enable critiquing?}
\label{sec_rq4}

\paragraph{Single-Step Critiquing.}\begin{figure}[t]
\centering
\begin{subfigure}[t]{0.4\textwidth}
\includegraphics[width=\textwidth]{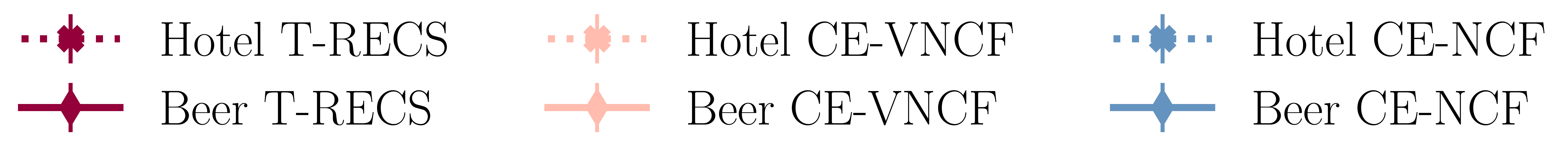}
\end{subfigure}
\begin{subfigure}[t]{0.20\textwidth}
    \centering
    \includegraphics[width=\textwidth,height=3.51cm]{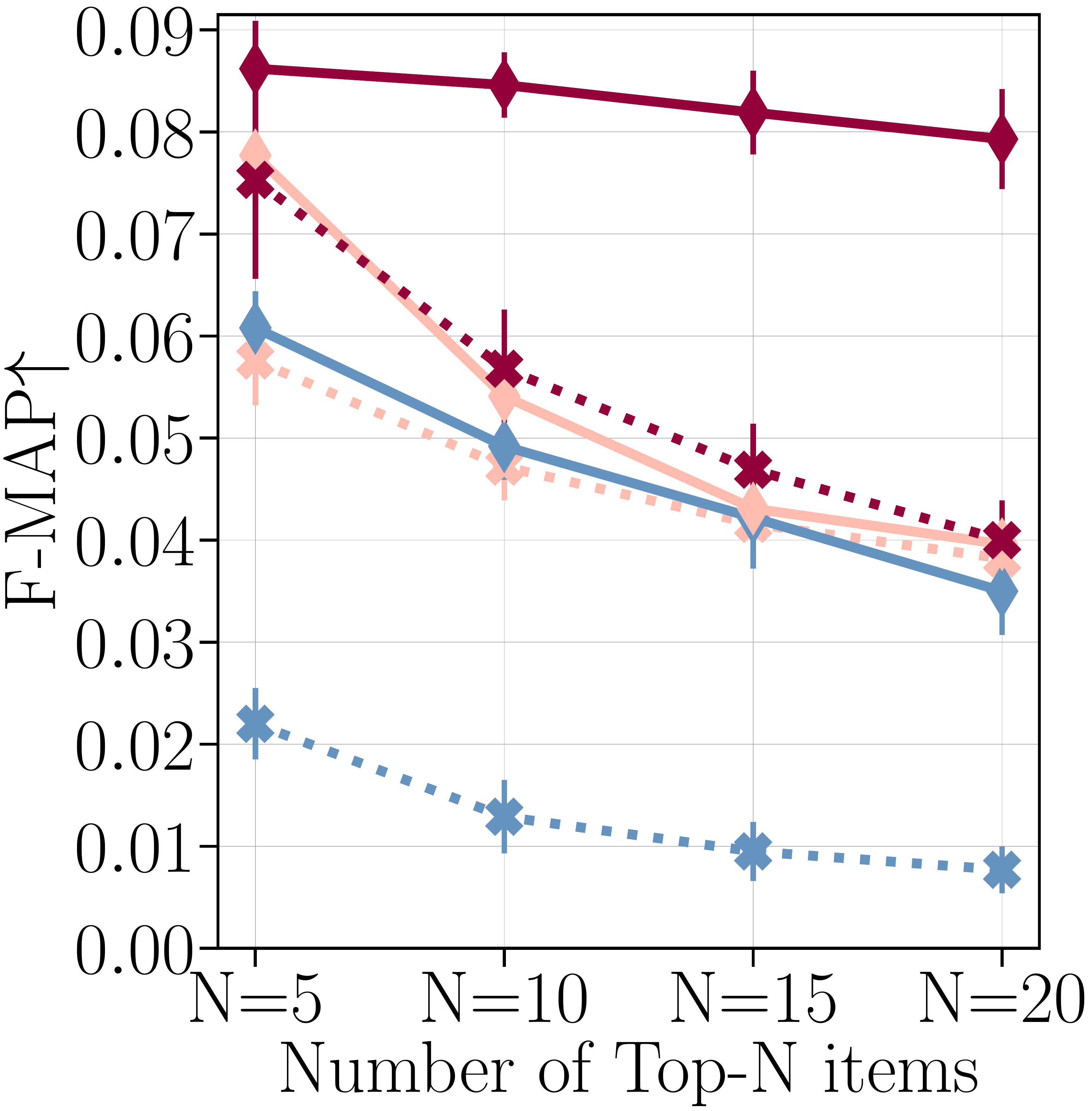}
    \caption{\label{single_crit_hotel}Falling MAP for different top-$N$. Error bars show the standard deviation.}
\end{subfigure}
\hfill
\begin{subfigure}[t]{0.275\textwidth}
    \centering
    \includegraphics[width=0.85\textwidth,height=3.51cm]{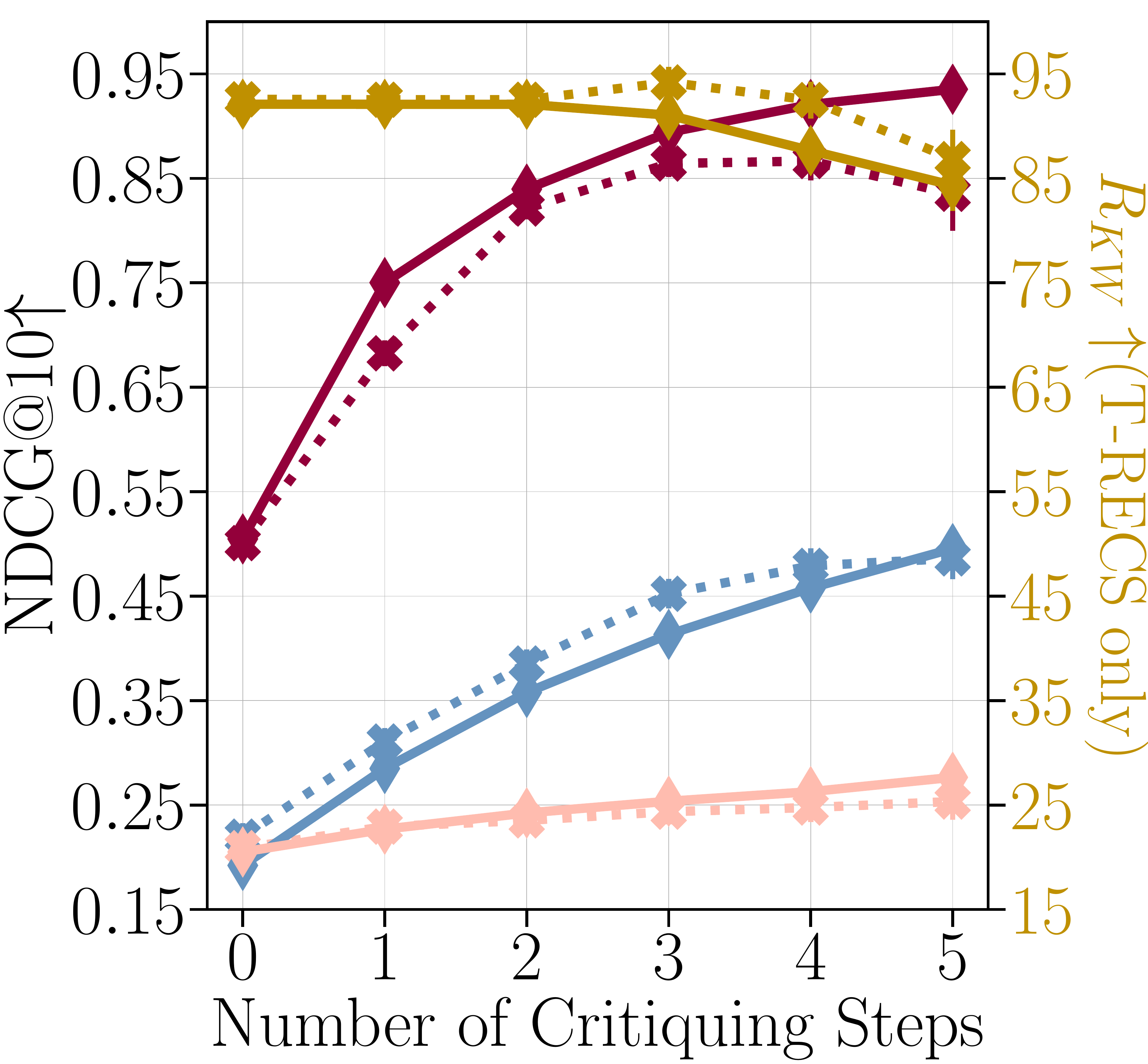}
    \caption{\label{mul_crit_hotel}Keyphrase prediction over multi-step critiquing with 95\% confidence interval. We also report the explanation consistency \textcolor{yellow2}{$R_{KW}$} for T-RECS.}
\end{subfigure}%
\caption{\label{single_mul_crit_both}Single- (top) and multi-step (bottom) critiquing.}
\end{figure}
For a given user, T-RECS recommends an item and generates personalized explanations, where the user can interact by critiquing one or multiple keyphrases. However, no explicit ground truth exists to evaluate the critiquing. We use F-MAP \cite{wu2019deep} to measure the effect of a critique. Given a user, a set of recommended items $\mathbb{S}$, and a critique~$k$, let $\mathbb{S}_k$ be the set of items containing $k$ in the explanation. The F-MAP measures the ranking difference of the affected items $\mathbb{S}_k$ before and after~critiquing $k$, using the Mean Average Precision at $N$. 
A positive F-MAP indicates that the rank of items in $\mathbb{S}_k$ fell after $k$ is critiqued. We compare T-RECS with CE-(V)NCF and average the F-MAP over $5,000$ user-keyphrase pairs.

Fig.~\ref{single_crit_hotel} presents the F-MAP performance on both datasets. All models show an anticipated positive F-MAP. The performance of T-RECS improves considerably on the beer dataset and is significantly higher for $N\le10$ on the hotel dataset. 
The gap in performance may be caused by the extra loss~of the autoencoder, which brings noise during training. T-RECS only iteratively edits the latent representation at test time.

\paragraph{Multi-Step Critiquing.}\label{sec_multi_crit}Evaluating multi-step critiquing via ranking is difficult because many items can have the keyphrases of the desired item. Instead, we evaluate whether a system obtains a complete model of the user's preferences following \cite{pu2006increasing}. A user expresses his keyphrase preferences iteratively according to a~randomly selected liked item. After each step, we evaluate the keyphrase explanations. For T-RECS, we also report the explanation consistency $R_{KW}$. 
We run up to five-steps critiques over $1,000$ random selected users and up to $5,000$ random keyphrases for each dataset. 
Fig.~\ref{mul_crit_hotel} shows that T-RECS builds through the critiques more accurate user profiles and consistent explanations. CE-NCF's top performance is significantly lower than T-RECS, and CE-VNCF plateaus, surely because of the KL divergence regularization, which limits the amount of information stored in the latent space. The explanation quality in T-RECS depends on the accuracy of the user's profile and may become saturated once we find it after four~steps.\footnote{We could not compare T-RECS with \cite{chen2020towards} because the authors did not make the code available due to copyright issues.}

\section{Conclusion}
Recommendations can carry much more impact if they are supported by explanations. 
We presented T-RECS, a multi-task learning Transformer-based recommender, that produces explanations considered significantly superior when evaluated by humans.
The second contribution of T-RECS is the user's ability to react to a recommendation by \textit{critiquing} the explanation. 
We designed an unsupervised method for multi-step critiquing with explanations. Experiments show that T-RECS obtains stable and significant improvement in adapting to the preferences expressed in multi-step critiquing.
\clearpage
\bibliographystyle{named}

\newpage
\clearpage
\appendix

\section{Processing \& Filtering \textit{Markers}}
\label{app_process_filtering}

The method described in \cite{antognini2019multi} extracts, most of the time, \textit{markers} that consist of long, continuous spans of words. However, sometimes, the \textit{markers} are too short because some reviews do not include enough words to justify a certain aspect rating, or the \textit{markers} stop in the middle of a sentence. Although both are theoretically not wrong, we aim to create fluent and grammatically correct justifications. To this end, we exploit the constituency parse tree to ensure that \textit{markers} are noun/verb phrases. We apply the following steps to the entire set of reviews for each dataset:

\begin{enumerate}
    \item Compute the constituency parse tree of each review;
    \item For each noun and adjective node in the constituency parse tree of a \textit{marker}, if the parent node is a verb or noun phrase, we add its children to the \textit{marker}. We follow the rules in \cite{giannakopoulos2017dataset};
    \item Filter out \textit{markers} having less than four tokens or including first and third-person pronouns.
\end{enumerate}
In cases where faceted ratings are not available at large cases, \cite{mukherjee2020uncertainty,niu-etal-2020-self} proposed elegant solutions to infer them from 20 or fewer samples. If faceted ratings are not present, one can use the concept rationalizer of \cite{antognini-2021-concept} to infer \textit{markers}.
\section{Keyphrase Samples}
\label{app_kws_datasets}

None of our datasets contains initially preselected keyphrases. We extract $200$ keyphrases from the \textit{markers} used to model the user and item profiles. They serve~as~a basis for the explanation and the critiquing. Table~\ref{kws_datasets} shows some keyphrases for each dataset. We apply the following processing for each~dataset, similarly to \cite{wu2019deep}:
\begin{enumerate}
    \item Group by aspect the \textit{markers} from all reviews. The aspect sets come from the available faceted ratings;\footnote{For the hotel reviews: service, cleanliness, value, location, and room. For beer reviews: appearance, smell, mouthfeel, and taste.}
    \item For each group of \textit{markers}:
    \begin{itemize}
        \item Tokenize and lemmatize the entire set of \textit{markers};
        \item Extract unigram lists of high-frequency noun and adjective phrases;
        \item Keep the top-$k$ most likely unigrams;
    \end{itemize}
    \item Represent each review as a one-hot vector indicating whether each keyphrase occurred in the review.
\end{enumerate}

Another possibility to extract keywords from reviews is to leverage Microsoft Concept Graph\footnote{\url{https://concept.research.microsoft.com/}} as in \cite{chen2019co,chen2020towards}. However, the provided API is limited to single instance conceptualization.
\begin{table}[h]
    \centering
   \caption{\label{kws_datasets}Some keyphrases mined from the inferred \textit{markers}. We grouped them by aspect for a better understanding.}
\begin{threeparttable}[t]
\begin{tabular}{@{}l@{\hspace{2mm}}l@{\hspace{2mm}}c@{}}
\textbf{Dataset} & \textbf{Aspect} & \textbf{Keyphrases}\\
\cmidrule[0.08em]{1-3}
\multirow{5}{*}{Hotel}
& Service & bar, lobby, housekeeping, guest\\
& Cleanliness & carpet, toilet, bedding, cigarette\\
& Value & price, wifi, quality, motel, gym\\
& Location & airport, downtown, restaurant shop\\
& Room & bed, tv, balcony, fridge, microwave\\
\cmidrule[0.08em]{1-3}
\multirow{4}{*}{Beer} & Appearance & golden, dark, white, foamy\\
& Aroma & fruit, wheat, citrus, coffee\\
& Palate & creamy, chewy, syrupy, heavy\\
& Taste & bitter, sweet, balanced, nutty\\
\cmidrule[0.08em]{1-3}
\end{tabular}
\end{threeparttable}
\end{table}

\section{Justification Examples}
\label{app_jus_ex}

Table~\ref{jus_comparison_hotel} and Table~\ref{jus_comparison_beer} present different justifications that are extracted from the hotel and beer reviews. We observe that the \textit{markers} justify the subratings. Although they might be some overlaps between EDU All and Markers, justifications from EDU All often are incomplete or not relevant.
\begin{table*}[!]
\footnotesize
    \centering
   \caption{\label{jus_comparison_hotel}Comparisons of the extracted justifications from different models for two hotels on the hotel dataset. Colors denote aspects while underline denotes EDUs classified as good justifications.}
\begin{threeparttable}
\begin{tabular}{@{}lm{6.8375cm}@{\hspace{6mm}}m{6.8375cm}@{}}
\textbf{Model} & \multicolumn{1}{c}{\textbf{Casa del Sol Machupicchu}} & \multicolumn{1}{c}{\textbf{Southern Sun Waterfront Cape Town}}\\
\toprule
Review & \underline{\textcolor{red}{the hotel was decent}} \textcolor{red}{ the staff was very friendly.} \underline{the free pisco} \underline{sour class with kevin was a nice bonus!} however, \textcolor{purple}{the rooms were lacking.} \textcolor{green}{the wifi was incredibly slow and there was no air conditioning, so it got very hot at night.} \textcolor{purple}{we couldn't open the windows either because there were so many bugs, birds, and noise.} \textcolor{blue}{overall, the location is convenient, but was is not worth the price.}\newline
& this is my second year visiting cape town and staying here. \underline{\textcolor{blue}{excellent location to business district, convention center, v\&a}} \underline{\textcolor{blue}{waterfront and access short distance to table mountain.}} \underline{\textcolor{red}{very nice}} \underline{\textcolor{red}{hotel, very friendly staff.} \textcolor{red}{breakfast is very good.}} \underline{\textcolor{purple}{rooms are nice}}~\textcolor{purple}{but bed mattress could be improved as bed is somewhat hard.} \underline{overall a very nice hotel.}\newline
\\
Rating & Overall: 3.0, \textcolor{red}{Service: 3.0}, \textcolor{yellow2}{Cleanliness: 4.0}, \textcolor{green}{Value: 2.0}, \textcolor{blue}{Location: 4.0}, \textcolor{purple}{Room: 3.0} & Overall: 4.0, \textcolor{red}{Service: 5.0}, \textcolor{yellow2}{Cleanliness: 5.0}, \textcolor{green}{Value: 4.0}, \textcolor{blue}{Location: 5.0}, \textcolor{purple}{Room: 3.0}\\
\midrule
Markers &
	-  \textcolor{purple}{the rooms were lacking.}\newline
	-  \textcolor{red}{the hotel was decent and the staff was very friendly.}\newline
	-  \textcolor{blue}{overall , the location is convenient , but was is not worth the price.}\newline
	-  \textcolor{purple}{we could n't open the windows either because there were so many bugs , birds , and noise.}\newline
	-  \textcolor{green}{the wifi was incredibly slow and there was no air conditioning , so it got very hot at night.}
	&
	-  \textcolor{red}{breakfast is very good.}\newline
	-  \textcolor{red}{very nice hotel , very friendly staff.}\newline
	-  \textcolor{purple}{rooms are nice but bed mattress could be improved as bed is somewhat hard.}\newline
	-  \textcolor{blue}{excellent location to business district , convention center , v\&a waterfront and access short distance to table mountain.}\\
\cdashlinelr{2-3}
EDU All &
	-  \underline{the hotel was decent}\newline
	-  \underline{the free pisco sour class with kevin was a nice bonus.}
&
	-  \underline{excellent location to business district , convention center , v\&a}\newline
	\  \underline{waterfront and access short distance to table mountain.}\newline
	-  \underline{very nice hotel , very friendly staff. breakfast is very good.}\newline
	-  \underline{rooms are nice.}\newline
	-  \underline{overall a very nice hotel.}\\
\cdashlinelr{2-3}
EDU One & -  the hotel was decent & -  very nice hotel , very friendly staff . breakfast is very good\\
\bottomrule
\end{tabular}
\end{threeparttable}
\end{table*}
\begin{table*}[!]
\footnotesize
    \centering
   \caption{\label{jus_comparison_beer}Comparisons of the extracted justifications from different models for two beers on the beer dataset. Colors denote aspects while underline denotes EDUs classified as good justifications.}
\begin{threeparttable}
\begin{tabular}{@{}lm{7.175cm}@{\hspace{6mm}}m{6.5cm}@{}}
\textbf{Model} & \multicolumn{1}{c}{\textbf{Saison De Lente}} & \multicolumn{1}{c}{\textbf{Bell's Porter}}\\
\toprule
Review
& poured from a 750ml bottle into a chimay branded chalice. a: \textcolor{red}{cloudy and unfiltered with a nice head that lasts and leaves good amounts of lacing in its tracks.} s: \textcolor{yellow2}{sour and bready with apple and yeast hints in there as well.} t: \textcolor{blue}{dry and hoppy with a nice crisp sour finish.} m: \underline{\textcolor{green}{medium bodied, high carbonation with big bubbles.}} d: \underline{easy to drink}, but i didn't really want more after splitting a 750ml with a buddy of mine.\newline
& \textcolor{red}{this beer pours black with a nice big frothy offwhite head.} \underline{\textcolor{yellow2}{smells or roasted malts, and chocolate.}} \underline{\textcolor{blue}{tastes of roasted malt with some chocolate and a hint of coffee.}} \textcolor{green}{the mouthfeel has medium body and is semi-smooth with some nice carbination.} \underline{drinkability is decent} i could drink a couple. \underline{overall a good choice} from bell's.\newline
\\
Rating & Overall: 3.0, \textcolor{red}{Appearance: 3.5}, \textcolor{yellow2}{Smell: 4.0}, \textcolor{green}{Mouthfeel: 3.5}, \textcolor{blue}{Taste: 3.5} & Overall: 3.5, \textcolor{red}{Appearance: 4.0}, \textcolor{yellow2}{Smell: 3.5}, \textcolor{green}{Mouthfeel: 3.5}, \textcolor{blue}{Taste: 4.0}\\
\midrule
Markers &
	- \textcolor{blue}{dry and hoppy with a nice crisp sour finish.}\newline
	- \textcolor{green}{medium bodied , high carbonation with big bubbles.}\newline
	- \textcolor{yellow2}{sour and bready with apple and yeast hints in there as~well.}\newline
	- \textcolor{red}{cloudy and unfiltered with a nice head that lasts and leaves good amounts of lacing in its tracks.}
&
	-  \textcolor{yellow2}{smells or roasted malts , and chocolate.}\newline
	-  \textcolor{red}{this beer pours black with a nice big frothy offwhite head.}\newline
	-  \textcolor{blue}{tastes of roasted malt with some chocolate and a hint of coffee.}\newline
	-  \textcolor{green}{the mouthfeel has medium body and is semi smooth with some nice carbination.}\\
\cdashlinelr{2-3}
EDU All &
- \underline{medium bodied , high carbonation with big bubbles.}\newline
- \underline{easy to drink} & 
	-  \underline{smells or roasted malts , and chocolate.}\newline
	-  \underline{tastes of roasted malt with some chocolate and a hint of~coffee.}\newline
	-  \underline{drinkability is decent}\newline
	-  \underline{overall a good choice}\\
\cdashlinelr{2-3}
EDU One & - easy to drink & -  drinkability is decent\\
\bottomrule
\end{tabular}
\end{threeparttable}
\end{table*}

\section{Full Natural Language Explanations Results}
\label{app_all_results}

We also compare T-RECS with more models than these of Section~\ref{sec_rq2}: LexRank~\cite{erkan2004lexrank}, NRT~\cite{li2017neural}, Item-Rand, Ref2Seq Top-k, and ACMLM \cite{ni-etal-2019-justifying}. LexRank \cite{erkan2004lexrank} is a unsupervised multi-document summarizer that selects an unpersonalized justification among all historical justifications of an item. NRT generates an explanation based on rating and the word distribution of the review. Item-Rand is an unpersonalized baseline which outputs a justification randomly from the justification history $J^i$ of item $i$. Ref2Seq Top-k is an extension of Ref2Seq, where we explore another decoding strategy called Top-k sampling \cite{radford2019language}, which should be more diverse and suitable on high-entropy tasks \cite{holtzman2019curious}. Finally, ACMLM is an aspect conditional masked language model that randomly chooses a justification from $J^i$ (similar to Item-Rand) and then iteratively edits it into new content by replacing random words. We also include more metrics and R\textsubscript{Sent}, which computes the percentages of generated justifications sharing the same polarity as the targets according to a sentiment classifier.\footnote{We employ the sentiment classifiers trained jointly with Multi-Target Masker of~\cite{antognini2019multi}, used to infer the \textit{markers} from which the justifications are extracted.}

The complete results are presented in Table~\ref{just_auto_perfs_add}. Interestingly, Item-Rand performs closely to LexRank: the best justification, according to LexRank, is slightly better than a random one. On the other hand, ACMLM edits the latter by randomly replacing tokens with the language model but produces poor quality justification, similarly to \cite{ni-etal-2019-justifying}. Finally, we also observe that the polarities of the generated justifications for beers match nearly perfectly, unlike in hotels where the positive and negative nuances are much harder to capture.
\begin{table*}[!]
\small
    \centering
\caption{\label{just_auto_perfs_add}Performance of the generated personalized justifications on automatic evaluation.}
\begin{threeparttable}
\begin{tabular}{@{}clcccc@{\hspace*{7mm}}ccc@{\hspace*{1mm}}c@{\hspace*{4mm}}c@{\hspace*{4mm}}cc@{}}
& \textbf{Model} & \textbf{B-1} & \textbf{B-2} & \textbf{B-3} & \textbf{B-4} & \textbf{R-1} & \textbf{R-2} & \textbf{R-L} & \textbf{BERT\textsubscript{Score}} & \textbf{PPL}$\downarrow$ & \textbf{R\textsubscript{KW}} & \textbf{R\textsubscript{Sent}}\\
\toprule
\multirow{11}{*}{\rotatebox{90}{\textit{Hotel}}} & Item-Rand & 11.50 & 2.88 & 0.91 & 0.32 & 12.65 & 0.87 & 9.75& 84.20 & - & 6.92 & 56.88\\
& LexRank & 12.12 & 3.31 & 1.10 & 0.41 & 14.74 & 1.16 & 10.61 & 83.91 & - & 10.32 & 58.51\\
& ExpansionNet & 4.03 & 1.95 & 1.01 & 0.53 & 34.22 & 9.65 & 6.91 & 74.81 & 28.87 & 60.09 & 61.38\\
& NRT & 17.00 & 6.51 & 3.06 & 1.51 & 18.21 & 2.88 & 16.08 & 86.31 & 29.75 & 11.44 & 64.46\\
& DualPC & 18.91 & 6.88 & 3.18 & 1.53 & 20.12 & 3.08 & 16.73 & 86.76 & 28.99 & 13.12 & 63.54\\
& CAML & 10.93 & 4.11 & 2.09 & 1.13 & 15.44 & 2.37 & 16.67 & 87.77 & 29.10 & 16.17 & 65.14\\
& Ref2Seq & 17.57 & 7.03 & 3.44 & 1.77 & 19.07 & 3.43 & 16.45 & 86.74 & \multirow{2}{*}{29.07} & 13.19 & 64.40\\
& Ref2Seq Top-k & 12.68 & 3.46 & 1.11 & 0.40 & 12.67 & 0.95 & 10.30 & 84.29 & & 6.38 & 58.11\\
& AP-Ref2Seq & 32.04 & 19.03 & 11.76 & 7.28 & 38.90 & 14.53 & 33.71 & 88.31 & 21.31 & 90.20 & 69.37\\
& ACMLM & 8.60 & 2.42 & 1.12 & 0.62 & 9.79 & 0.55 & 7.23 & 81.90 & - & 13.24 & 60.00\\
& T-RECS (Ours) & \textbf{33.53} & \textbf{19.76} & \textbf{12.14} & \textbf{7.47} & \textbf{40.29} & \textbf{14.74} & \textbf{34.10} & \textbf{90.23} & \textbf{17.80} & \textbf{93.57} & \textbf{70.12}\\
\bottomrule
\multirow{11}{*}{\rotatebox{90}{\textit{Beer}}} &  Item-Rand & 10.96 & 3.02 & 0.91 & 0.29 & 10.28 & 0.75 & 8.25 & 83.39 & - & 6.70 & 99.61\\
& LexRank & 12.23 & 3.58 & 1.12 & 0.38 & 13.81 & 1.16 & 9.90 & 83.42 & - & 10.79 & 99.88\\
& ExpansionNet & 6.48 & 3.59 & 2.06 & 1.22 & \textbf{54.53} & 18.24 & 9.68 & 72.32 & 22.28 & 82.49 & \textbf{99.99}\\
& NRT & 18.54 & 8.53 & 4.40 & 2.43 & 17.46 & 3.61 & 15.56 &  85.26 & 21.22 & 11.43 & \textbf{99.99}\\
& DualPC & 18.38 & 8.10 & 3.95 & 2.08 & 17.61 & 3.38 & 14.68 & 85.49 &21.15 & 10.60 & \textbf{99.99}\\
& CAML & 12.94 & 6.5 & 3.80 & 2.43 & 14.63 & 2.43 & 14.99 & 85.96 & 21.29 & 10.18 & \textbf{99.99}\\
& Ref2Seq & 18.75 & 9.47 & 5.51 & 3.51 & 18.25 & 4.52 & 15.96 & 85.27 & \multirow{2}{*}{22.34} & 12.10 & \textbf{99.99}\\
& Ref2Seq Top-k & 13.92 & 5.02 & 2.10 & 1.01 & 12.36 & 1.50 & 10.52 & 84.14 & & 8.51 & 99.83\\
& AP-Ref2Seq & 44.84 & 30.57 & 21.68 & 15.89 & 51.38 & 23.27 & 46.50 & 91.35 & 12.07 & 91.52 & \textbf{99.99}\\
& ACMLM & 7.76 & 2.54 & 0.91 & 0.34 & 8.33 & 0.87 & 6.17 & 80.94 & - & 10.33 & \textbf{99.99}\\
& T-RECS (Ours) & \textbf{46.50} & \textbf{31.56} & \textbf{22.42} & \textbf{16.54} & 53.12 & \textbf{23.86} & \textbf{47.20} & \textbf{91.50} & \textbf{10.24} & \textbf{94.96} & \textbf{99.99}\\
\bottomrule
\end{tabular}
\end{threeparttable}
\end{table*}
\begin{table*}[!]
\small
    \centering
\caption{\label{kw_perfs2}Performance of personalized keyphrase explanation quality.}
\begin{threeparttable}
\begin{tabular}{@{}ll@{\hspace{1.5mm}}c@{\hspace{1mm}}c@{\hspace{1mm}}cc@{\hspace{1.5mm}}c@{\hspace{1mm}}c@{\hspace{1mm}}cc@{\hspace{1.5mm}}c@{\hspace{1mm}}c@{\hspace{1mm}}cc@{\hspace{1.5mm}}c@{\hspace{1mm}}c@{\hspace{1mm}}c@{}}
& & \multicolumn{3}{c}{\textbf{NDCG@N}} & & \multicolumn{3}{c}{\textbf{MAP@N}} & & \multicolumn{3}{c}{\textbf{Precision@N}} & & \multicolumn{3}{c}{\textbf{Recall@N}}\\
\cmidrule{3-5}\cmidrule{7-9}\cmidrule{11-13}\cmidrule{15-17}
& \textbf{Model} & N=5 & N=10 & N=20 & & N=5 & N=10 & N=20 & & N=5 & N=10 & N=20 & & N=5 & N=10 & N=20\\
\toprule
\multirow{8}{*}{\rotatebox{90}{\textit{Hotel}}} & UserPop & 0.2625 & 0.3128 & 0.3581 & & 0.2383 & 0.1950 & 0.1501 & & 0.1890 & 0.1332 & 0.0892 & & 0.2658 & 0.3694 & 0.4886\\
& ItemPop & 0.2801 & 0.3333 & 0.3822 & & 0.2533 & 0.2083 & 0.1608 & & 0.2041 & 0.1431 & 0.0959 & & 0.2866 & 0.3961 & 0.5245\\
\cdashlinelr{2-17}
& E-NCF & 0.2901 & 0.3410 & 0.3889 & & 0.2746 & 0.2146 & 0.1618 & & 0.1943 & 0.1366 & 0.0919 & & 0.2746 & 0.3802 & 0.5057\\
& CE-NCF & 0.1929 & 0.2286 & 0.2634 & & 0.1825 & 0.1432 & 0.1085 & & 0.1290 & 0.0918 & 0.0631 & & 0.1809 & 0.2548 & 0.3469\\
\cdashlinelr{2-17}
& E-VNCF & 0.2902 & 0.3441 & 0.3925 & & 0.2746 & 0.2158 & 0.1634 & & 0.1947 & 0.1391 & 0.0932 & & 0.2746 & 0.3860 & 0.5132\\
& CE-VNCF & 0.1727 & 0.2289 & 0.2761 & & 0.1530 & 0.1336 & 0.1115 & & 0.1275 & 0.1071 & 0.0767 & & 0.1795 & 0.2965 & 0.4200\\
\cdashlinelr{2-17}
& T-RECS (Ours) & \textbf{0.3158} & \textbf{0.3763} & \textbf{0.4319} & & \textbf{0.2919} & \textbf{0.2356} & \textbf{0.1807} & & \textbf{0.2223} & \textbf{0.1581} & \textbf{0.1068} & & \textbf{0.3109} & \textbf{0.4358} & \textbf{0.5812}\\
\cmidrule[0.125em]{1-17}
\multirow{8}{*}{\rotatebox{90}{\textit{Beer}}} & UserPop & 0.2049 & 0.2679 & 0.3357 & & 0.2749 & 0.2404 & 0.2014 & & 0.2366 & 0.1901 & 0.1445 & & 0.1716 & 0.2767 & 0.4207\\
& ItemPop & 0.1948 & 0.2495 & 0.3131 & & 0.2653 & 0.2291 & 0.1894 & & 0.2267 & 0.1759 & 0.1342 & & 0.1618 & 0.2529 & 0.3886\\
\cdashlinelr{2-17}
& E-NCF & 0.1860 & 0.2485 & 0.3158 & & 0.2488 & 0.2204 & 0.1877 & & 0.2162 & 0.1789 & 0.1389 & & 0.1571 & 0.2618 & 0.4040\\
& CE-NCF & 0.1471 & 0.1922 & 0.2422 & & 0.1967 & 0.1721 & 0.1446 & & 0.1687 & 0.1363 & 0.1050 & & 0.1227 & 0.1974 & 0.3033\\
\cdashlinelr{2-17}
& E-VNCF & 0.1763 & 0.2362 & 0.3055 & & 0.2389 & 0.2097 & 0.1797 & & 0.2031 & 0.1696 & 0.1356 & & 0.1471 & 0.2478 & 0.3943\\
& CE-VNCF & 0.1512 & 0.2025 & 0.2595 & & 0.1987 & 0.1784 & 0.1532 & & 0.1774 & 0.1475 & 0.1155 & & 0.1293 & 0.2146 & 0.3352\\
\cdashlinelr{2-17}
& T-RECS (Ours) & \textbf{0.2394} & \textbf{0.3163} & \textbf{0.3946} & & \textbf{0.3127} & \textbf{0.2799} & \textbf{0.2369} & & \textbf{0.2800} & \textbf{0.2284} & \textbf{0.1717} & & \textbf{0.2048} & \textbf{0.3320} & \textbf{0.4970}\\
\bottomrule
\end{tabular}
\end{threeparttable}
\end{table*}

\section{Full Keyphrase Explanation Results}
Table~\ref{kw_perfs2} contains complementary results to the keyphrase explanation quality experiment of Section \ref{sec_keyphrases_explanations}.

\section{Additional Metrics Multi-Step Critiquing}
\label{app_add_multistep}
\begin{figure}[!t]
\centering
\begin{subfigure}[t]{0.5\textwidth}
    \centering
    \includegraphics[width=\textwidth]{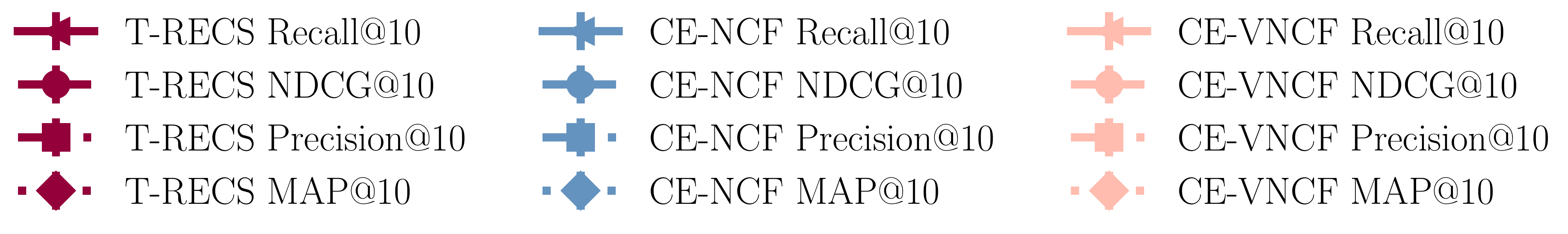}
\end{subfigure}
\begin{subfigure}[t]{0.35\textwidth}
    \centering
    \includegraphics[width=\textwidth]{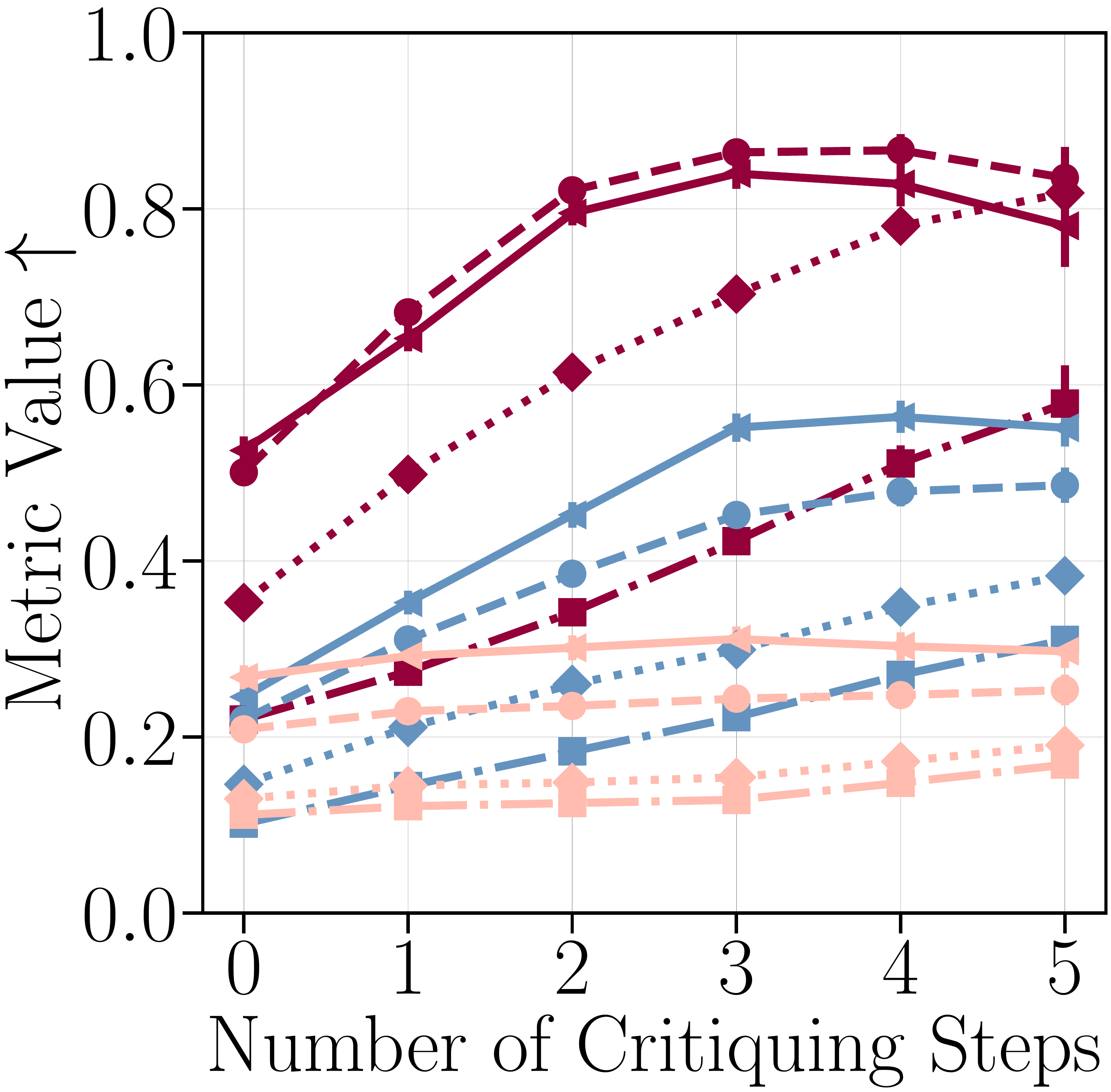}
    \caption{Results on the \textbf{hotel} dataset.}
\end{subfigure}
\hspace{0.1\textwidth}
\begin{subfigure}[t]{0.35\textwidth}
    \centering
    \includegraphics[width=\textwidth]{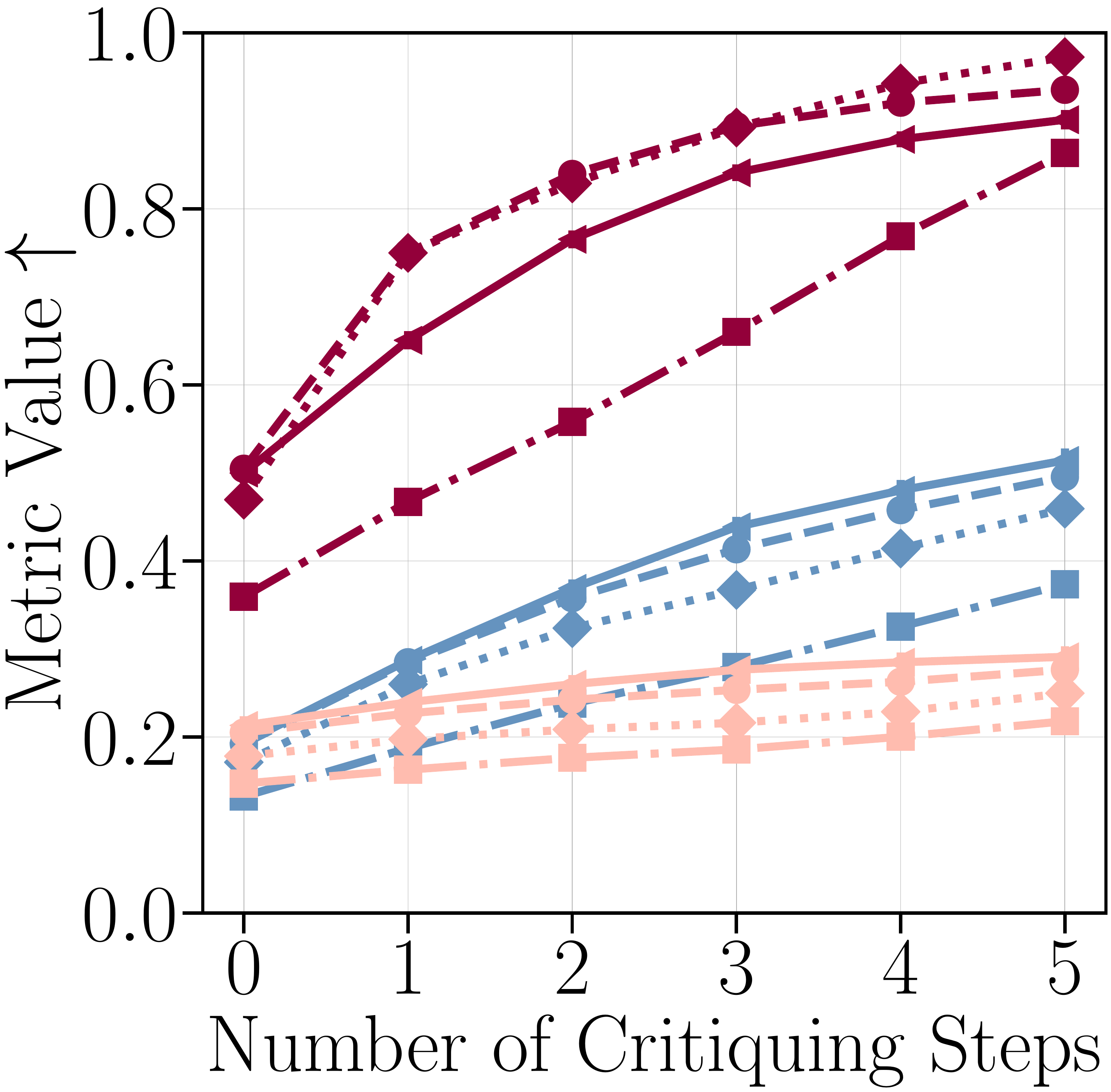}
    \caption{Results on the \textbf{beer} dataset.}
\end{subfigure}
\caption{\label{mul_mul_crit_both}Multi-step critiquing performance. Keyphrase prediction over multi-step critiquing in terms of Recall@10, NDCG@10, Precision@10, and MAP@10 with 95\% confidence interval. a)~Results on the hotel dataset, b)~on the beer dataset.}
\end{figure}

More metrics of the multi-step critiquing experiment in Section~\ref{sec_multi_crit} are available in Fig.~\ref{mul_mul_crit_both}.

\section{RQ 4: Do T-RECS justifications benefit the overall recommendation quality?}
\label{sec_rq3}In this section, we investigate whether justifications are beneficial to T-RECS and improve overall performance. We assess the performance on three different axes: rating prediction, preference prediction, and Top-N recommendation.

\subsubsection{Rating \& Preference Prediction}
\label{sec_rating}
\begin{table}[t]
\small
    \centering
   \caption{\label{rat_perfs}Performance of the rating prediction.}
\begin{threeparttable}
\begin{tabular}{@{}l@{\hspace{1.5mm}}c@{\hspace{1.5mm}}c@{\hspace{1.5mm}}c@{}l@{\hspace{4mm}}c@{\hspace{1.5mm}}c@{\hspace{1.5mm}}c@{}}
& \multicolumn{3}{c}{\textit{Hotel}} & & \multicolumn{3}{c}{\textit{Beer}}\\
\cmidrule{2-4}\cmidrule{6-8}
\textbf{Model} & \textbf{MAE} & \textbf{RMSE} & \textbf{$\tau \uparrow$} & & \textbf{MAE} & \textbf{RMSE} & \textbf{$\tau \uparrow$}\\
\toprule
NMF & 0.3825 & 0.6171 & 0.2026 & & 0.3885 & 0.4459 & 0.4152\\
PMF & 0.3860 & 0.5855 & 0.0761 & & 0.3922 & 0.4512 & 0.4023\\
HFT & 0.3659 & 0.4515 & 0.4584 & & 0.3616 & 0.4358 & 0.4773\\
NARRE & 0.3564 & 0.4431 & 0.4476 & & 0.3620 & 0.4377 & 0.4506\\
\cdashlinelr{1-8}
NCF & 0.3619 & 0.4358 & 0.4200 & & 0.3638 & 0.4341 & 0.4696\\
E-NCF & 0.3579 & 0.4382 & 0.4145 & & 0.3691 & 0.4326 & 0.4685\\
CE-NCF & 0.3552 & 0.4389 & 0.4165 & & 0.3663 & 0.4390 & 0.4527\\
\cdashlinelr{1-8}
VNCF & 0.3502 & 0.4313 & 0.4408 & & 0.3666 & 0.4300 & 0.4706\\
E-VNCF & 0.3494 & 0.4365 & 0.4072 & & 0.3627 & 0.4457 & 0.4651\\
CE-VNCF & 0.3566 & 0.4545 & 0.3502 & & \textbf{0.3614} & 0.4330 & 0.4619\\
\cdashlinelr{1-8}
T-RECS & \textbf{0.3306} & \textbf{0.4305} & \textbf{0.4702} & & \textbf{0.3614} & \textbf{0.4295} & \textbf{0.4909}\\
\bottomrule
\end{tabular}
\end{threeparttable}
\end{table}
We first analyze recommendation performance by the mean of rating prediction. We utilize the common Mean Squared Error (MSE) and Root Mean Squared Error (RMSE) metrics. However, the rating prediction performance alone does not best reflect the quality of recommendations, because users mainly see the relative ranking of different items~\cite{ricci2011introduction,musat2015personalizing}. Consequently, we measure also how well the item rankings computed by T-RECS agree with the user's own rankings as given by his own review ratings. We measure this quality by leveraging the standard metric Kendall's $\tau$ rank correlation~\cite{kendall1938measure}, computed on all pairs of rated-items by a user in the testing set. Overall, there are $153\,954$ and $1\,769\,421$ pairs for the hotel and beer datasets, respectively.

We examine the following baseline methods together with T-RECS: NMF~\cite{hoyer2004non} is a non-negative matrix factorization model for rating prediction. PMF~\cite{mnih2008probabilistic} is a probabilistic matrix factorization method using ratings for collaborative filtering. HFT~\cite{mcauley2013hidden} is a strong latent-factor baseline, combined with a topic model aiming to find topics in the review text that correlate with the users' and items' latent factors. NARRE~\cite{chen2018neural} is a state-of-the-art model that predicts ratings and reviews' usefulness jointly. Finally, we include the six methods of~\cite{wu2019deep} described in Section~\ref{sec_keyphrases_explanations}.

The results are shown in Table~\ref{rat_perfs}. T-RECS consistently outperforms all the baselines, by a wide margin on the hotel dataset, including models based on collaborative filtering with/without review information or models extended with an explanation component and/or a critiquing component. Interestingly, the improvement in the hotel dataset in terms of MAE and RMSE is significantly higher than on the beer dataset. We hypothesize that this behavior is due to the~sparsity (see Table~\ref{stats_datasets}), which has also been observed in the hotel domain in prior work~\cite{musat2015personalizing,antognini_hotel_rec}. On the beer dataset, we~note that reviews contain strong indicators and considerably improve the performance of NARRE and HFT compared to collaborative filtering methods. The extended (V)NCF models with either an explanation and/or a critiquing component improve MAE performance. Therefore, explanations can benefit the recommender systems to improve rating~prediction.

Table~\ref{rat_perfs} also contains the results in terms of preference prediction. T-RECS achieves up to $0.0136$ higher Kendall~correlation compared to the best baseline. Surprisingly, we note that CE-VNCF, NMF, and PMF show much worse results on the hotel datasets than on the beer dataset. This highlights that hotel reviews are noisier than~beer reviews and emphasizes the importance of capturing users' profiles, where T-RECS does best in comparison to other~models.

\subsubsection{Preference Prediction}
\begin{figure}[!t]
\centering
\begin{subfigure}[t]{0.4\textwidth}
    \centering
    \includegraphics[width=\textwidth]{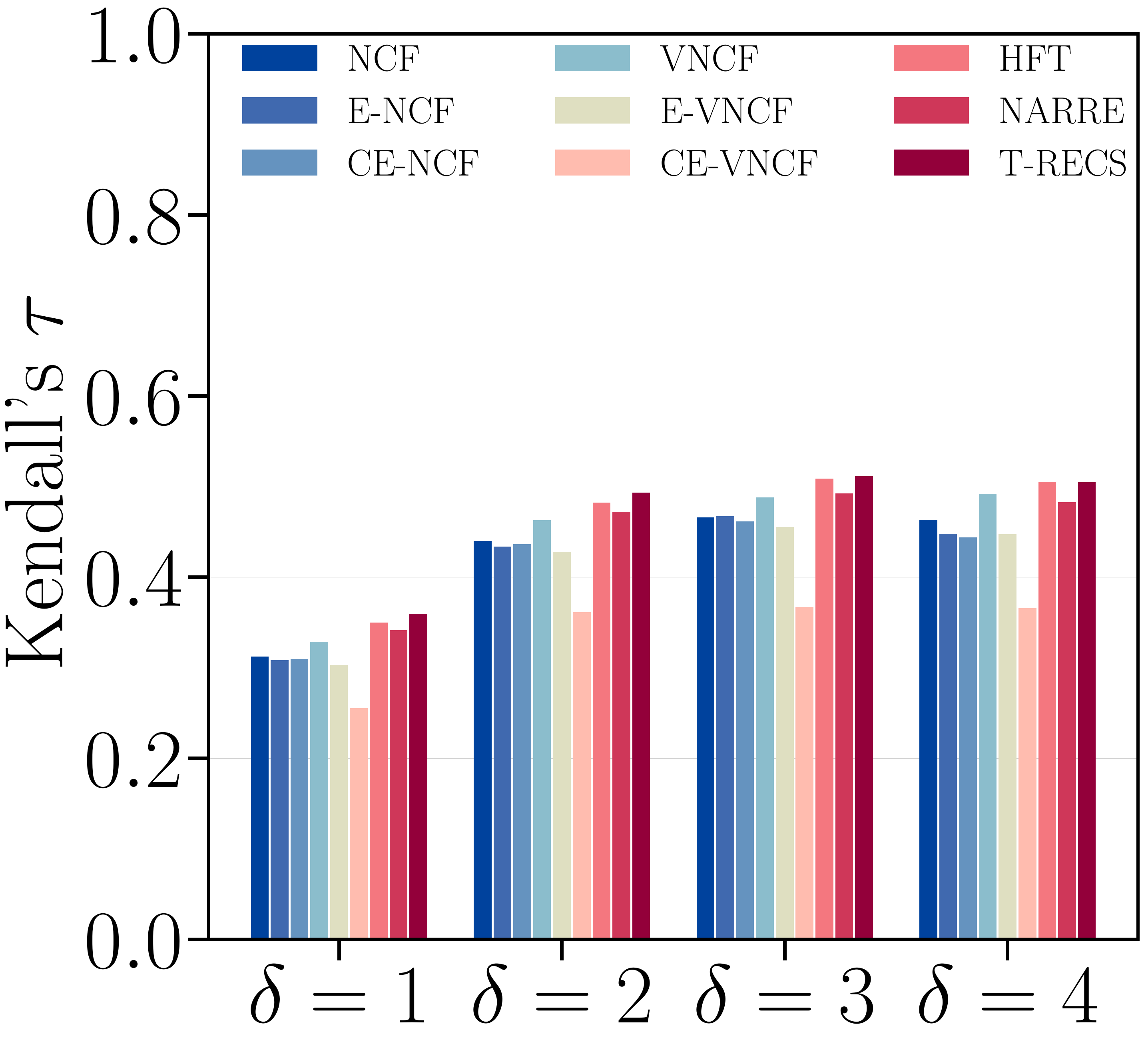}
    \caption{\label{kendall_hotel}Kendall's $\tau$ correlation on the \textbf{hotel} dataset.}
\end{subfigure}%
\hfill
\begin{subfigure}[t]{0.4\textwidth}
    \centering
    \includegraphics[width=\textwidth]{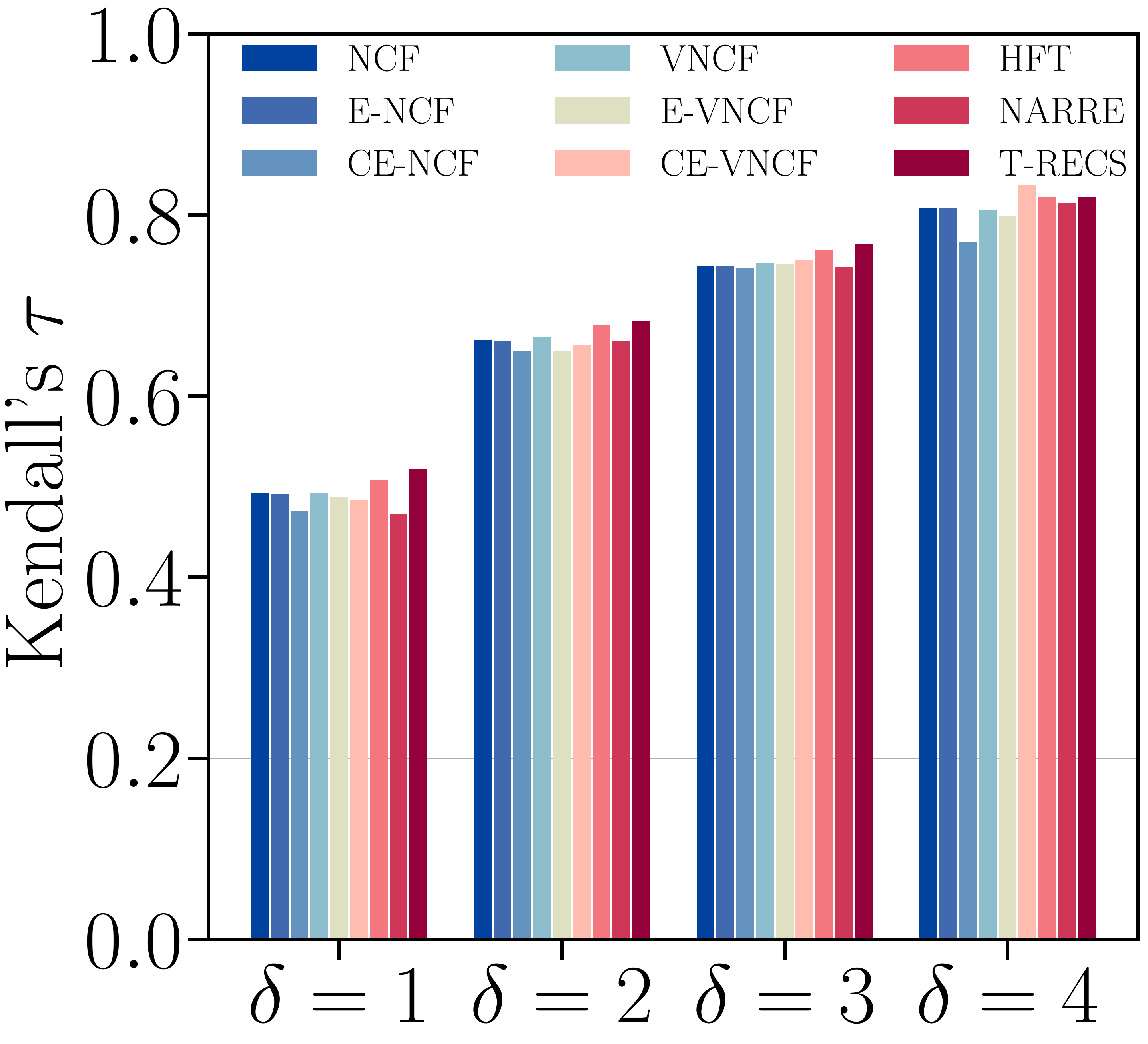}
    \caption{\label{kendall_beer}Kendall's $\tau$ correlation on the \textbf{beer} dataset.}
\end{subfigure}
\caption{\label{kendal_both}Performance of the preference prediction using Kendall's $\tau$ and $\delta = |y^i_r  - y^j_r|$.}
\end{figure}
In this experiment, we study a more fine-grained rank correlation. Following \cite{musat2015personalizing}, we analyze the pairwise ranking of rated items by a user, and we impose a minimum value for the rating difference between two items $i$ and $j$, such that $\delta = |y^i_r  - y^j_r|$; the rating difference $\delta$ symbolizes the minimum preference strength. 

Fig.~\ref{kendal_both} contains the Kendall's $\tau$ evaluation for multiple $\delta$ on both datasets. Overall, T-RECS increases the Kendall correlation similarly to other models but performs better on average. We observe that HFT's performance is similar to T-RECS, although slightly lower for most cases. On the beer dataset, we surprisingly note that CE-VNCF obtains a negligible higher score for $\delta = 4$, while significantly underperforming for $\delta < 4$, and especially on the hotel dataset. Finally, the Kendall's $\tau$ correlation increases majorly with the strength of preference pairs on the beer dataset and plateaus over $\delta \ge 2$ on the hotel dataset. It highlights that hotel reviews are noisier than beer reviews, and it emphasizes the importance of capturing users' profiles, where T-RECS does best in comparison to other models.
\begin{table}[!t]
\small
    \centering
\caption{\label{rec_perfs}Performance of the Top-$N$ recommendation.}
\begin{threeparttable}[ht]
\begin{tabular}{@{}ll@{\hspace{1mm}}c@{\hspace{2mm}}c@{}l@{\hspace{3mm}}c@{\hspace{1mm}}c@{}l@{\hspace{1mm}}c@{\hspace{2mm}}c@{}}
& & \multicolumn{2}{c}{\textbf{NDCG@N}} & & \multicolumn{2}{c}{\textbf{Precision@N}} & & \multicolumn{2}{c}{\textbf{Recall@N}}\\
\cmidrule{3-4}\cmidrule{6-7}\cmidrule{9-10}
& \textbf{Model} & N=10 & N=20 & & N=10 & N=20 & & N=10 & N=20\\
\toprule
\multirow{8}{*}{\rotatebox{90}{\textit{Hotel}}} & NCF & 0.1590 & 0.2461 & & 0.0231 & 0.0200 & & 0.2310 & 0.3991\\
& E-NCF & 0.1579 & 0.2432 & & 0.0234 & 0.0200 & & 0.2336 & 0.4004\\
& CE-NCF & 0.1585 & 0.2431 & & 0.0235 & 0.0201 & & 0.2352 & 0.4028\\
\cdashlinelr{2-10}
& VNCF & 0.1492 & 0.2431 & & 0.0220 & 0.0197 & & 0.2204 & 0.3932\\
& E-VNCF & 0.1505 & 0.2395 & & 0.0219 & 0.0192 & & 0.2188 & 0.3842\\
& CE-VNCF & \textbf{0.1738} & \textbf{0.2662} & &0.0221 & 0.0190 & & 0.2210 & 0.3809\\
\cdashlinelr{2-10}
& T-RECS & 0.1674 & \textbf{0.2662} & & \textbf{0.0236} & \textbf{0.0207} & & \textbf{0.2358} & \textbf{0.4144} \\
\cmidrule[0.125em]{1-10}
\multirow{8}{*}{\rotatebox{90}{\textit{Beer}}} & NCF & 0.2172 & 0.3509 & & 0.0250 & 0.0212 & &  0.2499 & 0.4231 \\
& E-NCF & 0.2087 & 0.3363 & & 0.0243 & 0.0205 & & 0.2426 & 0.4103\\
& CE-NCF & 0.2226 & 0.3456 & & 0.0252 & 0.0205 & & 0.2517 & 0.4105\\
\cdashlinelr{2-10}
& VNCF & 0.1943 & 0.3329 & & 0.0235 & 0.0211 & & 0.2345 & 0.4213\\
& E-VNCF & 0.1387 & 0.2813 & & 0.0158 & 0.0168 & & 0.1579 & 0.3362\\
& CE-VNCF & 0.2295 & 0.3598 & & 0.0263 & 0.0218 & & 0.2630 & 0.4352\\
\cdashlinelr{2-10}
& T-RECS & \textbf{0.2372} & \textbf{0.3674} & & \textbf{0.0269} & \textbf{0.0219} & &  \textbf{0.2693} & \textbf{0.4390}\\
\bottomrule
\end{tabular}
\end{threeparttable}
\end{table}

\subsubsection{Recommendation Performance}
We evaluate the performance of T-RECS on the last dimension: Top-N recommendation. We adopt the widely used leave-one-out evaluation protocol~\cite{PERDIF2019,zhao2018categorical}; in particular, for each user, we randomly select one liked item in the test set alongside 99 randomly selected unseen items. We compare T-RECS with the state-of-the-art methods in \cite{wu2019deep}. Finally, we rank the item lists based on the recommendation scores produced by each method, and report the NDCG, Precision, and Recall at different~N.

Table~\ref{rec_perfs} presents the main results. 
Comparing to CE-(V)NCF models, which contain an explanation and critiquing components, our proposed model shows better recommendation performance for almost all metrics on the two datasets. On average, the variants of (V)NCF reach higher results than the original method, which was not the case in the rating prediction and relative rankings tasks (see Section~\ref{sec_rating}), unlike T-RECS that shows consistent results.

\section{Human Evaluation Details}
\label{app_hes}

We use Amazon's Mechanical Turk crowdsourcing platform to recruit human annotators to evaluate the quality of extracted justifications and the generated justifications produced by each model. To ensure high-quality of the collected data, we restricted the pool to native English speakers from the U.S., U.K., Canada, or Austria. Additionally, we set the worker requirements at a 98\% approval rate and more than 1000 HITS.

\begin{figure*}[]
\centering
\includegraphics[width=0.75\linewidth]{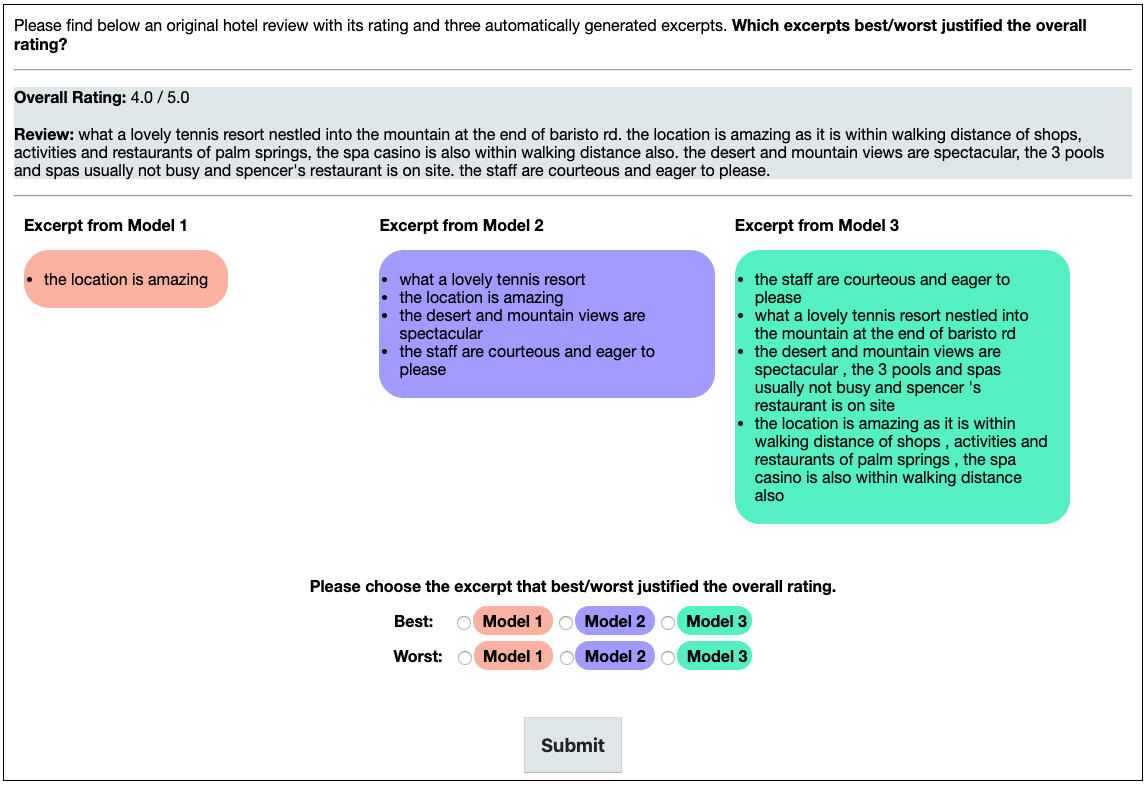}
\caption{\label{he_int_ex_ju}Annotation platform for judging the quality of extracted justifications from different methods. The justifications are shown in random order for each comparison. In this example, our method corresponds to the third model.}
\end{figure*}

\begin{figure*}[]
\centering
\includegraphics[width=0.75\linewidth]{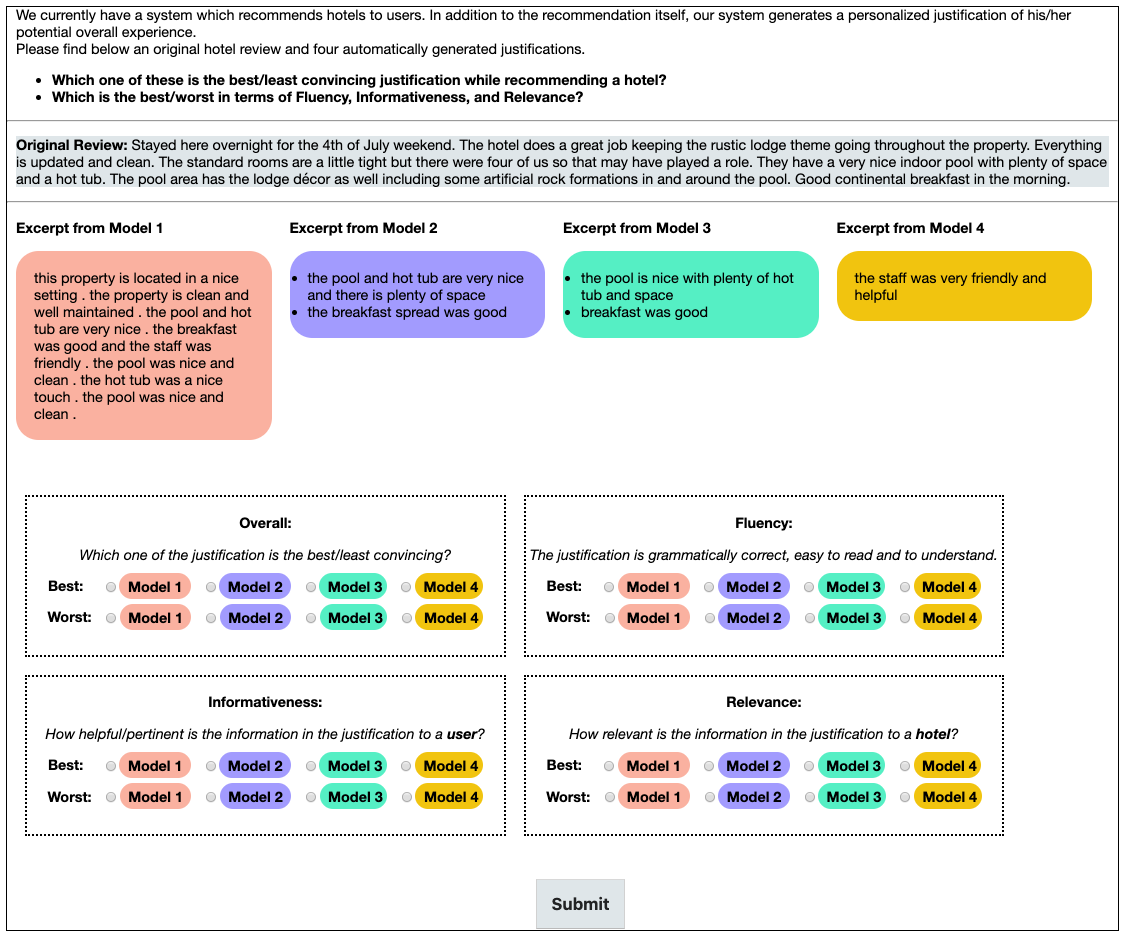}
\caption{\label{he_int_ge_ju}Annotation platform for judging the quality of generated justifications from different methods, on four dimensions. The justifications are shown in random order for each comparison. In this example, our method corresponds to the second~model.}
\end{figure*}

The user interface used to judge the quality of the justifications extracted from different methods, in Section~\ref{sec_rq1}, is shown in Fig.~\ref{he_int_ex_ju}. Another human assessment evaluates the generated justifications (see Section~\ref{sec_rq2}) on the four dimensions: \begin{enumerate*}
    \item \underline{overall} measures the overall subjective quality;
    \item \underline{fluency} represents the readability;
    \item \underline{informativeness} indicates whether the justification contains information pertinent to the user;
    \item \underline{relevance} measures how relevant the information is to an item
\end{enumerate*}. The interface is available in Fig.~\ref{he_int_ge_ju}

\section{Additional Training Details}
\label{app_reproducibility}

\subsection{Tuning}
We build the justification history $J^u$,$J^i$, with $N_{just}=32$. In T-RECS, we set the embedding, latent, and self-attention dimension size to 256, and the dimension of the~feed-forward network to 1024. The encoder and decoder consist of two layers of Transformer with $4$~attention heads. We use a batch size of 128, dropout of 0.1, 4000 warm-up steps, smoothing parameter $\varepsilon=0.1$, and Adam with learning rate $0.001,\beta_1=0.9,\beta_2=0.98$, and $\epsilon=10^{-9}$. The~Rating Classifier and Keyphrase Explainer are two layers of 128 and 64 dimensions with LeakyReLU ($\alpha=0.2$).~For critiquing, we choose a threshold and decay coefficient $T=0.015,\zeta=0.9$ and $T=0.01,\zeta=0.975$~for hotel and beer reviews, respectively. We use the code from the authors for most models. We tune all models on the dev set. We have operated a random search over $10$ trials. We chose the models achieving the lowest validation loss. The range of hyperparameters are the following for T-RECS (similar for other models):

\begin{itemize}
	\item Learning rate: $[0.001, 0.0001]$;
	\item Max epochs: $[100, 200, 300]$;
	\item Batch size: $[128]$;
	\item Hidden size encoder/decoder: $[256]$;
	\item Attention heads: $[4]$;
	\item Number of layers: $[2]$;
	\item Dropout encoder: $[0.0, 0.1, 0.2, 0.3, 0.4, 0.5]$;
	\item Dropout decoder: $[0.0, 0.1, 0.2, 0.3]$;
	\item General dropout: $[0.0, 0.1, 0.2]$;
	\item Warmup: $[2000, 4000, 8000, 16000]$;
	\item $\lambda_{r}$, $\lambda_{kp}$, $\lambda_{just}$: $[1.0]$;
\end{itemize}

Most of the time, the model converges under $20$ epochs.
For critiquing, we employed:
\begin{itemize}
	\item Decay coefficient $\zeta$: $[0.5, 0.75, 0.8, 0.9, 0.95]$;
	\item Max iterations: $[25, 50, 75, 100, 200]$;
	\item Threshold: $[0.015, 0.01, 0.005]$;
\end{itemize}

\subsection{Hardware / Software}

\begin{itemize}
	\item \textbf{CPU}: 2x Intel Xeon E5-2680 v3 (Haswell), 2x 12 cores, 24 threads, 2.5 GHz, 30 MB cache;
	\item \textbf{RAM}: 16x 16GB DDR4-2133;
	\item \textbf{GPU}: 2x Nvidia Titan X Maxwell;
	\item \textbf{OS}: Ubuntu 18.04;
	\item \textbf{Software}: Python 3.6, PyTorch 1.3.0, CUDA 10.0.
\end{itemize}

\subsection{Running Time}

Currently, our code is not optimized, and the critiquing experiment has been run on CPU. In this setting, the critique of a user takes less than 2.5 seconds. By optimizing the code, batching critiques together, and leveraging GPUs/TPUs, we would expect a significant gain. In practice, we could also limit the critique to the Top-N items (e.g., 100 or 1000). To keep track of the critiques and user representations, one could use a database to avoid recomputing the previous critiques' representations.

\subsection{Addressing Users without Reviews}

The cold-start problem is not particular to our method but a general problem in recommendation. However, our system could infer $\bm{\gamma^u}$ and $\bm{\beta^u}$ for users without reviews and then computes the initial latent representation $\bm{z}$ as in Section~\ref{sec_model}. We propose the following strategies:
\begin{enumerate}
	\item \textbf{Users with ratings but no reviews}: we could leverage collaborative filtering techniques to identify users with similar ratings and build an initial representation (a similar method is used in \cite{multi_step_demo}).
	\item \textbf{New users}: following the observation of \cite{zhang2014explicit,musat2015personalizing}: “users write opinions about the topics they care about” (mentioned in Section~\ref{sec_extr_just}), we could ask new users to write about the different aspects they deem important. Their initial representation is an aggregation of other users with similar writing. Another option is to ask the users to select items they like based on a subset of items and build a profile from these preferences (see above).
\end{enumerate}

\end{document}